\documentclass[preprint,12pt]{elsarticle}

\usepackage{amssymb}
\usepackage{amsmath}
\usepackage{tabularx}
\usepackage{array}
\usepackage{booktabs} % opcional pero recomendable
\usepackage{threeparttable}
\usepackage{makecell}
\usepackage{color}
\usepackage{hyperref}
\usepackage{xurl}
\usepackage{soul}
\usepackage[normalem]{ulem}
\newcolumntype{L}[1]{>{\raggedright\arraybackslash}p{#1}}
\newcolumntype{C}[1]{>{\centering\arraybackslash}p{#1}}
\newcolumntype{R}[1]{>{\raggedleft\arraybackslash}p{#1}}
\newcolumntype{Y}{>{\raggedright\arraybackslash}X}

\AtBeginEnvironment{table}{\normalsize\renewcommand{\arraystretch}{1.15}\setlength{\tabcolsep}{4pt}}
\AtBeginEnvironment{table*}{\normalsize\renewcommand{\arraystretch}{1.15}\setlength{\tabcolsep}{4pt}}

\setcellgapes{2pt}
\makegapedcells

\usepackage{graphicx}     % usually loaded by elsarticle, but safe
\usepackage{subcaption}   % \begin{subfigure} ... \end{subfigure}
\usepackage{url}
\usepackage{amsmath}
\usepackage{amssymb}
\newtoggle{finalPaper}
\toggletrue{finalPaper} % Change \toggletrue <-> \togglefalse

\iftoggle{finalPaper} {
    \newcommand{\addtxt}[1]{#1}
    
    \newcommand{\change}[2]{#2}
    \newcommand{\rmvtxt}[1]{}
} {
    \newcommand{\addtxt}[1]{\textcolor{blue}{#1}}
    
    \newcommand{\change}[2]{\sout{#1}\textcolor{blue}{#2}}
    \newcommand{\rmvtxt}[1]{\sout{#1}}
}

\journal{Applied Soft Computing}

\begin{document}

\begin{frontmatter}



\title{
SYNAPSE: Framework for Neuron Analysis and Perturbation in Sequence Encoding}


\author[1]{Jes\'us S\'anchez Ochoa}\ead{jesus.sanchezo@um.es}
\author[1]{Enrique Tom\'as Mart\'inez Beltr\'an\corref{cor1}}\ead{enriquetomas@um.es}
\author[1]{Alberto Huertas Celdr\'an}\ead{alberto.huertas@um.es}

\affiliation[1]{organization={Department of Information and Communications Engineering, University of Murcia},
            city={Murcia},
            postcode={30100}, 
            country={Spain}}
\cortext[cor1]{Corresponding author.}

%% Abstract
\begin{abstract}
In recent years, Artificial Intelligence has become a powerful partner for complex tasks such as data analysis, prediction, and problem-solving, yet its lack of transparency raises concerns about its reliability. In sensitive domains such as healthcare or cybersecurity, ensuring transparency, trustworthiness, and robustness is essential, since the consequences of wrong decisions or successful attacks can be severe. Prior neuron-level interpretability approaches are primarily descriptive, task-dependent, or require retraining, which limits their use as systematic, reusable tools for evaluating internal robustness across architectures and domains. To overcome these limitations, this work proposes SYNAPSE, a systematic, training-free framework (requiring no retraining of the base model) for understanding and stress-testing the internal behavior of Transformer models across domains. SYNAPSE extracts per-layer \texttt{[CLS]} representations, trains a lightweight linear probe to obtain global and per-class neuron rankings, and applies forward-hook interventions during inference. This design enables controlled, repeatable experiments on internal representations without altering the original model, thereby allowing weaknesses, stability patterns, and label-specific sensitivities to be measured and compared directly across tasks and architectures. Across all experiments, SYNAPSE reveals a \change{consistent, domain-independent organization of internal representations}{consistent organization of internal representations across the two evaluated domains}, in which task-relevant information is encoded in broad, overlapping neuron subsets. This redundancy provides a strong degree of functional stability, while class-wise asymmetries expose heterogeneous specialization patterns and enable fine-grained, label-aware analysis. In contrast, small structured manipulations in weight or logit space are sufficient to redirect predictions, highlighting complementary vulnerability profiles and illustrating how SYNAPSE can guide the development of more robust and interpretable Transformer models.

\end{abstract}


% \begin{highlights}
%     \item SYNAPSE is a training-free framework for neuron-level interpretability and robustness analysis in Transformer classifiers.
%     \item Linear probes rank neurons globally and per class using layer-wise [CLS] representations.
%     \item Non-destructive forward-hook interventions enable targeted ablations during inference without retraining.
%     \item Cross-domain experiments show task information is widely distributed across overlapping neuron subsets.
%     \item Small weight/logit edits redirect predictions more easily than neuron silencing, exposing asymmetric robustness.
% \end{highlights}

%% Keywords
\begin{keyword}
Interpretability \sep Robustness \sep Privacy \sep Model Auditing \sep Representation learning \sep Model Evaluation \sep Explainability 


\end{keyword}

\end{frontmatter}



\section{Introduction}

With the rise of AI comes serious concerns about its reliability, trustworthiness, and robustness. As models become more complex, they become black boxes, making it difficult to understand how they reach their conclusions. In critical scenarios, such as medical diagnosis, military anti-missile systems, or malware detection, the inability to explain a decision can have catastrophic consequences. Explainable AI (XAI) seeks to shed light on internal processes and provide human-interpretable explanations, a crucial task as complexity grows and transparency is traded for high predictive performance \cite{Rudin2019}. While researchers have developed techniques to support validation and enhance robustness, most progress remains focused on Natural Language Processing (NLP) \cite{arrieta,Danilevsky2020}. Furthermore, increasing regulations and social concerns, such as the EU AI Act, drive interest in avoiding bias and ensuring fair interactions \cite{EUAIAct2024}.
In recent years, advances in RNNs and LSTMs have enabled high-level performance, yet they face limitations in parallelization and in capturing long-range dependencies \cite{Cho2014}. To overcome these, the Transformer architecture and its self-attention mechanism seek to determine which parts of the input data should be held in consideration the most, giving more importance (or paying more attention) to the most relevant elements in a sequence \cite{vaswani2017attention}. This led to the development of Large Language Models (LLMs) such as GPT, BERT, and RoBERTa \cite{Brown2020,Devlin2019,Liu2019}. Its impact, however, extends beyond language: the same self-attention principle underpins vision Transformers for medical image segmentation \cite{myovitsnet}. These models, pre-trained on massive datasets, can be fine-tuned for diverse tasks with relatively little data \cite{Devlin2019,Liu2019}. Recently, this has been applied to cybersecurity, with studies showing that LLMs can achieve state-of-the-art results in malware detection using system call sequences \cite{sanchez2024transfer}. Recent surveys on LLMs for cyber-resilience further emphasize the gap between external attack methods and internal model-auditing tools~\cite{harnessingLLMcyber}, a gap this work aims to address.
Despite the growing body of work on neuron-level analysis, several limitations remain. Current approaches based on the representation erasure measure assess the impact of removing parts of the internal signal, but they operate as broad perturbation methods and do not provide a fine-grained mechanism to localize and manipulate individual functional units in a controlled, reusable manner \cite{li2017understanding}. Subsequent work demonstrated that specific neurons can be identified and even modified to influence model behavior; however, these methods typically rely on task-specific training procedures or architecture-dependent pipelines, which limit their applicability as general, training-free analysis tools \cite{bau2019identifying}. Other studies have shown that high-level behaviors such as sentiment can emerge in individual units of generative models, yet these findings are largely observational and do not provide a systematic framework to measure robustness or to perform targeted interventions without affecting the rest of the network \cite{radford2017learning}. More recently, empirical evidence suggests that task-relevant knowledge is often distributed across many neurons rather than isolated in a small subset, making it difficult to assess how internal representations contribute to model decisions using simple attribution or ablation strategies \cite{song2024neurons}. In addition, most neuron-level interpretability advances have been developed and evaluated in the context of Natural Language Processing, with comparatively little effort devoted to designing domain-independent methodologies that can be transferred across tasks and data modalities.

To address this gap, this work shifts from descriptive neuron analysis to an interventional and operational perspective, in which internal units are systematically ranked and intervened at inference time to quantify their functional role. In this setting, neuron-level interpretability is not treated as a purely explanatory tool but as an experimental mechanism to assess robustness, sensitivity, and class-conditional behavior across architectures and domains. This objective is articulated through the following research questions:

\begin{enumerate}[RQ1]
\item Global sensitivity. How many top-$k$ neurons must be silenced to impact overall performance across models and domains significantly?
\item Label-aware brittleness. Does silencing a small, label-aware subset of neurons meaningfully affect metrics tied to a target label?
\item Class-conditional vulnerability. Which labels are most sensitive to per-class neuron silencing, and does this pattern differ between malware and language domains?
\end{enumerate}

To operationalize this interventional, training-free analysis setting, this work introduces SYNAPSE, a framework for systematically analyzing and stress-testing the internal behavior of Transformer-based models across domains. Rather than providing purely descriptive explanations, SYNAPSE enables controlled, inference-time interventions that quantify the functional role of internal units in terms of robustness, sensitivity, and class-conditional behavior. Concretely, the contributions of this work are as follows:

\begin{itemize}

\item \textbf{SYNAPSE framework.}
A modular and non-destructive pipeline that automatically extracts layer-wise \texttt{[CLS]} activations, trains lightweight linear probes to obtain global and class-conditional neuron importance rankings, and performs targeted interventions through forward hooks without retraining.

\item \textbf{Intervention-based silencing strategies for neuron-level robustness evaluation.}
Three complementary intervention mechanisms are introduced: 
(i) \textit{global undirected silencing}, which removes the top-$k$ neurons according to a global ranking to measure overall sensitivity; 
(ii) \textit{global directed silencing}, which selects globally important neurons with maximal influence on a target label to perform label-aware analysis; and 
(iii) \textit{per-class silencing}, which suppresses neurons associated with a specific class to directly probe class-conditional brittleness.

\item \textbf{\change{Efficient and architecture-agnostic analysis}{Efficient analysis across the evaluated Transformer architectures}.}
The use of compact \texttt{[CLS]} representations enables a computationally efficient and scalable neuron-level pipeline applicable to different Transformer encoders, while a mean-pooled representation extends the same pipeline to decoder-only models such as GPT-2, so that the analysis spans both encoder-only and decoder-only architectures. \addtxt{The evaluation demonstrates the applicability of the workflow to four encoder checkpoints and one decoder-only checkpoint, using architecture-appropriate sequence representations.}

\item \textbf{Cross-domain experimental validation.}
A unified evaluation protocol spanning malware detection from system-call sequences and emotion classification from natural language, allowing direct comparison of neuron-level behavior across heterogeneous modalities.

\end{itemize}

To our knowledge, SYNAPSE is the first training-free, reversible framework to unify probe-based neuron-importance ranking with inference-time, forward-hook interventions into a single reusable protocol that operates consistently across encoder-only and decoder-only Transformer architectures and across heterogeneous domains, here malware detection and emotion classification, and that quantifies internal robustness through comparable failure budgets, namely how much targeted intervention is required before performance degrades.

The remainder of this paper is organized to reflect the development of the proposed approach. Section \ref{sec:rw} positions this work within the literature on neuron-level interpretability and adversarial analysis. Section \ref{sec:tm} introduces the threat model that defines the realistic conditions under which training-free internal interventions are carried out. The design and components of the SYNAPSE framework are presented in Section \ref{sec:fw}. The experimental protocol, including datasets, evaluation metrics, and architectural settings, is detailed in Section \ref{sec:val}. The results obtained across domains are reported and analyzed in Section \ref{sec:results}. Finally, Section \ref{sec:disc} discusses the main findings and limitations, and Section \ref{sec:conc} concludes the paper and outlines future research directions.

\section{Related Work}\label{sec:rw}

This section reviews recent work along two main research directions: (i) neuron-level explainable AI methods, which aim to identify and analyze neurons that encode class-specific information, and (ii) attacks on machine learning models, which study how targeted manipulations can induce misclassification. Table~\ref{table:neuron_analysis} and Table~\ref{table:ml_attacks} provide a comparative taxonomy of representative methods across these two axes and serve as a structural reference for the subsequent subsections.

\subsection{Explainable Artificial Intelligence}

A trend in XAI techniques is to perform intervention-based neuron analysis of the model, which involves actively intervening on neurons to test their effects on model behavior. The goal of these techniques is to identify important neurons by ablating or manipulating them and observing changes in their outputs, an approach adopted in this work. Li et al. \cite{li2017understanding} used this idea to understand neural network representations, removing neurons from RNNs to see the impact on translation accuracy, confirming that neurons indeed carried specific linguistic information. Afterwards, Bau et al. \cite{bau2019identifying} expanded on this idea by identifying and directly controlling key neurons in a neural machine translation model, showing that activating or suppressing these neurons could modify specific behaviors. A different approach was presented in Radford et al. \cite{radford2017learning}, which sought a single ``sentiment neuron'' that could alter the sentiment of generated text by adjusting its activation. Lastly, Song et al. \cite{song2024neurons} focused on neurons in LLMs that may be uniquely responsible for certain tasks and found that disabling them reduced task performance, confirming their causal role in those tasks. Together, these studies demonstrate that individual neurons can assume salient, interpretable roles. However, they also reveal that scaling to larger architectures (e.g., LLMs) requires more systematic methods for locating and manipulating these salient units, a gap this work aims to fill.

Another line of work studies how individual neurons specialize in representing specific concepts within neural models. Early findings show that internal representations tend to organize into functional substructures, with different neurons capturing distinct linguistic properties such as sentiment, syntax, or semantics.

This idea is formalized through Linguistic Correlation Analysis, where simple probe models are trained on individual neuron activations to predict interpretable features and quantify neuron specialization. This approach was introduced by Dalvi et al. \cite{dalvi2018grain} and later extended by Durrani et al. \cite{durrani2020analyzing}. Building on this methodology, Dalvi et al. also released NeuroX, a toolkit that provides modular support for neuron-level probing and remains one of the few frameworks explicitly designed to operate at this level of granularity \cite{dalvi2023neurox}.

\begin{table*}[!t]
\centering
\caption{Comparison of Neuron Analysis Methods.}
\label{table:neuron_analysis}

\scriptsize
\renewcommand{\arraystretch}{1.15}
\setlength{\tabcolsep}{3.5pt}

\begin{tabularx}{\textwidth}{
  >{\centering\arraybackslash}p{0.9cm}  % Reference (fixed)
  >{\centering\arraybackslash}p{0.9cm}    % Year (fixed)
  >{\raggedright\arraybackslash}p{2.2cm}  % Model Type (fixed)
  >{\raggedright\arraybackslash}X         % Domain / Purpose (flex)
  >{\raggedright\arraybackslash}X         % Explanation Type (flex)
  >{\raggedright\arraybackslash}p{1.6cm}  % Scope (fixed)
}
\hline
\textbf{Ref.} & \textbf{Year} & \textbf{Model Type} & \textbf{Domain / Purpose} & \textbf{Explanation Type} & \textbf{Scope} \\
\hline
\cite{li2017understanding} & 2017 & LSTM-based NLP classifier & NLP (sentiment, linguistic tasks) & Feature ablation & \makecell[l]{Local\\Global} \\
\cite{radford2017learning} & 2017 & mLSTM & NLP (unsupervised text generation) & Emergent feature analysis & Global \\
\cite{bau2019identifying} & 2019 & NMT & NLP (machine translation) & Neuron importance ranking and manipulation & Global \\
\cite{dalvi2018grain} & 2018 & Pre-trained RNN & NLP (language tasks) & Neuron interpretation & Global \\
\cite{durrani2020analyzing} & 2020 & Transformer (BERT) & NLP (language understanding) & Neuron importance and clustering & Global \\
\cite{dalvi2023neurox} & 2023 & Transformers & NLP (interpretability) & Toolkit supporting multiple neuron analysis methods & \makecell[l]{Local\\Global} \\
\cite{song2024neurons} & 2024 & LLM & NLP (multi-task) & Causal analysis & Global \\
\hline
\end{tabularx}
\end{table*}

\subsection{Attacks on Machine Learning Models}

In general, white-box evasion techniques assume the attacker has access to the full model, including gradients, training data, and model weights, making it easier to craft highly effective adversarial examples that can significantly degrade model precision. Firstly, Szegedy et al. \cite{szegedy2014intriguingpropertiesneuralnetworks} demonstrated that imperceptible, L-BFGS-optimized perturbations revealed deep networks' vulnerability to carefully crafted adversarial examples. Building on this, several authors have examined the impact of imperceptible input perturbations on misclassification in deep networks, as in Goodfellow et al. \cite{goodfellow2015explainingharnessingadversarialexamples}, who introduced FGSM (Fast Gradient Sign Method). This method uses the gradient to craft perturbations that maximize loss in a quick one-step attack. Further research by Pravin et al. \cite{pravin2021adversarial} proposed a neuron-fragility-exploitation attack that identifies and perturbs a small set of highly sensitive neurons in ResNet models to induce misclassifications, an approach that motivates this work. Another important contribution was made by Che et al. \cite{Che}, who extended white-box attacks to LLMs by manipulating latent and weight-space features. This precise category was chosen when this work was developed, as robustness can be measured with no extra effort when the model's critical data are available.

\begin{table*}[!t]
\centering
\caption{Adversarial and poisoning attacks in machine learning.}
\label{table:ml_attacks}

\scriptsize
\renewcommand{\arraystretch}{1.15}
\setlength{\tabcolsep}{3.0pt}

\begin{tabularx}{\textwidth}{
  >{\centering\arraybackslash}p{0.9cm} % Reference
  >{\centering\arraybackslash}p{0.9cm}   % Year
  >{\raggedright\arraybackslash}p{1.3cm} % Threat Model
  >{\raggedright\arraybackslash}X        % Attack Type
  >{\raggedright\arraybackslash}p{1.4cm} % Phase
  >{\raggedright\arraybackslash}p{1.5cm} % Access
  >{\raggedright\arraybackslash}p{1.3cm} % Model Type
}
\hline
\textbf{Ref.} & \textbf{Year} & \textbf{Threat Model} & \textbf{Attack Type} & \textbf{Phase} & \textbf{Access} & \textbf{Model Type} \\
\hline
\cite{szegedy2014intriguingpropertiesneuralnetworks} & 2014 & Evasion & Small perturbation adversarial examples (L-BFGS method) & Inference & White-box & DNN (CNN) \\
\cite{goodfellow2015explainingharnessingadversarialexamples} & 2015 & Evasion & FGSM (Fast Gradient Sign Method) & Inference & White-box & DNN (CNN) \\
\cite{pravin2021adversarial} & 2021 & Evasion & Neuron-fragility exploitation (target-specific ``fragile'' neurons) & Inference & White-box & DNN (ResNet) \\
\cite{Che} & 2024 & Capability elicitation & Hidden feature/weight manipulation (latent and weight-space attacks) & Inference (fine-tune) & White-box & LLM (GPT) \\
\hline
\end{tabularx}
\end{table*}

In conclusion, the reviewed literature shows that neuron-level analysis and adversarial robustness have largely evolved along separate paths. Existing interpretability methods successfully identify neurons with causal or specialized roles, but they are predominantly descriptive, strongly tied to NLP tasks, and lack a systematic and reusable protocol for large Transformer architectures. Furthermore, most adversarial approaches focus on input-space perturbations or direct weight manipulation, providing limited insight into how internal representations contribute to model stability or vulnerability. Moreover, current techniques rarely enable controlled, training-free interventions that can be consistently applied across domains and models. As a result, there is still no unified framework that enables the internal sources of robustness and fragility to be located, ranked, and experimentally tested in a \change{comparable, architecture-agnostic manner}{manner that remains comparable across models of differing depth, attention pattern, and context length}.

\section{Threat Model} \label{sec:tm}

In the context of deploying LLMs in high-stakes domains such as cybersecurity and defense, this work considers a threat landscape in which models face not only input-level perturbations but also internal manipulation and operational degradation. The necessity for the SYNAPSE framework is motivated by several realistic threat scenarios that directly correspond to the neuron intervention strategies developed:

\begin{itemize}

    \item \textbf{TH-1: Traffic evasion by mimicking legitimate behavior.} In real networks, attackers frequently adapt malicious communication to resemble normal traffic, for example, by blending command-and-control exchanges into common patterns that appear to be regular HTTPS or DNS traffic. The objective is straightforward: reduce the observable signals that security systems use to discriminate between malicious and benign activity, thereby increasing the probability that harmful traffic is accepted as normal.

    \item \textbf{TH-2: Compromised or altered model before deployment.} A separate and increasingly common risk is that the model itself is modified somewhere along the deployment pipeline, such as during training, packaging, storage, or distribution. In this scenario, the model appears to work correctly in routine use, but behaves incorrectly under attacker-chosen conditions, for instance, misclassifying a specific type of malicious activity while leaving other predictions largely unaffected. This kind of tampering can be subtle, difficult to detect, and operationally attractive because it preserves an appearance of normal performance. In the experiments conducted in this work, this scenario is emulated by temporarily and reversibly editing the classification head at inference time, which isolates the behavioral effect of such tampering without persistently modifying the deployed model.
    
    \item \textbf{TH-3: Faults that corrupt computation during inference.} Deployed systems may experience transient faults that do not change the input data but still perturb the internal computation, such as memory bit flips, hardware instability, or electromagnetic interference in edge or critical environments. Even when rare, these events can distort intermediate representations and lead to unreliable or unsafe decisions, particularly in settings where the classifier is part of an automated response pipeline.
    
    \item \textbf{TH-4: Manipulation of the serving process and final outputs.} If the inference service is compromised, an attacker may not need to alter inputs or model weights to influence decisions. Instead, they can directly steer predictions at the output stage, for example, by biasing scores toward the normal class or by altering the decision logic applied to logits. This scenario is particularly relevant in production deployments, where the model runs as a service and its outputs are consumed downstream by monitoring or mitigation components.

\end{itemize}

To make the correspondence between the threat scenarios and the experiments explicit, each intervention instantiates a specific scenario: input-space perturbations (random noise injection and FGSM) instantiate TH-1; weight-space edits of the classification head (class-directed output reweighting, balanced weight push, class suppression, and the bias-only variant) instantiate TH-2, emulated reversibly at inference time; neuron silencing (global and per-class), Gaussian noise injection, and the attribution-guided bit-flip instantiate TH-3; and logit-bias manipulation instantiates TH-4. This mapping is recalled in the corresponding result subsections.

\section{Synapse Framework} \label{sec:fw}
\begin{figure*}[!t]
    \centering
    \includegraphics[width=\textwidth]{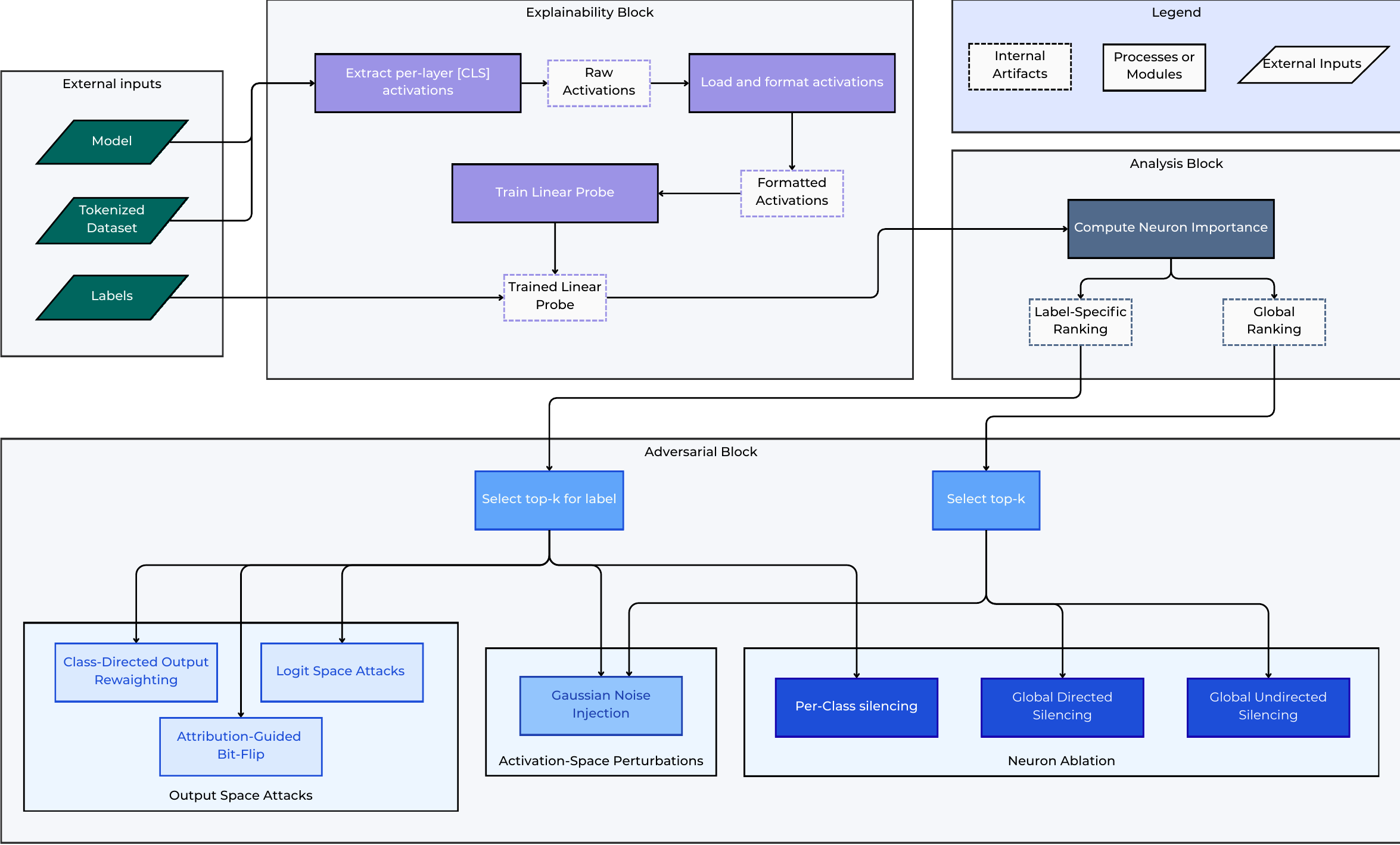}
    \caption{Internal structure of SYNAPSE}
    \label{fig:design}
\end{figure*}

This section details the system design used to study neuron-level behavior and robustness. The framework comprises three blocks that orchestrate the analysis and intervention process: (i) Explainability Block, which extracts \texttt{[CLS]} activations layer-wise (i.e., the sequence-level representation used for classification) and trains a lightweight linear probe to quantify neuron importance; (ii) Analysis Block, which converts probe weights into global and per-class rankings, mapping top-$k$ neuron indices to specific layers and hidden dimensions to enable reproducible, scope-controlled interventions, carried out in the (iii) Adversarial Block. In addition, it implements a variety of inference-time perturbations via PyTorch forward hooks, including global undirected, label-directed, and per-class silencing, as well as auxiliary stress tests such as logit bias, Gaussian noise, and weight-tilting. This modular, training-free, and reversible setup is illustrated in Figure~\ref{fig:design}, which shows the complete architecture of the framework.

This work is inspired by the approach developed in Dalvi et al. \cite{dalvi2023neurox} and their NeuroX library, which served as the basis for activation extraction, probe training, and neuron importance computation. Nevertheless, these tools had to be adapted to meet the requirements of this work.

The framework operates as a cyclic workflow to ensure experimental integrity and reproducibility. Each execution follows a six-step protocol:
\begin{enumerate}
    \item \textbf{Ranking:} Compute global and class-specific rankings on a held-out data split.
    \item \textbf{Selection:} Identify the top-$k$ indices based on the desired $p$ and scope.
    \item \textbf{Intervention:} Register PyTorch forward hooks to intercept and modify activations during inference.
    \item \textbf{Inference:} Run the model on the test set and record performance metrics.
    \item \textbf{Cleanup:} Remove all active hooks to restore the model to its original state.
    \item \textbf{Verification:} Recompute the baseline to ensure no permanent changes were made to the weights.
\end{enumerate}

This non-destructive intervention is made possible by forward hooks. In the PyTorch ecosystem, hooks are user-defined mechanisms registered with modules or tensors to enable the inspection, extraction, or modification of intermediate computations. They are triggered at specific stages of execution (such as the forward or backward pass) without altering the model's underlying architecture. This allows the framework to dynamically register and unregister modifications as needed, ensuring that no permanent changes are made to the model's weights. In this work, forward hooks are used to silence targeted neurons during inference, enabling stress testing of the model's robustness in a reversible setup.

All parameters, including random seeds and the specific JSON-persisted neuron indices, are logged to guarantee that any intervention can be exactly replicated across both the malware and GoEmotions domains.

\subsection{Formalization}
Let a Transformer encoder with $L$ layers and hidden size $H$ map an input sequence of $T$ tokens to per-layer hidden states $h^{(l)}_t \in \mathbb{R}^{H}$. SYNAPSE summarizes each layer by either the classification token, $h^{(l)}_{\texttt{[CLS]}}$, or, for architectures without such a token, by a mean-pooled representation
\begin{equation}\bar{h}^{(l)} = \frac{1}{T}\sum_{t=1}^{T} h^{(l)}_t .\end{equation}
Stacking the retained layers (excluding the embedding and final classification layers) yields a feature vector $\phi(x)\in\mathbb{R}^{D}$ with $D = L\cdot H$.

A lightweight linear probe with parameters $W_p\in\mathbb{R}^{C\times D}$ and $b_p\in\mathbb{R}^{C}$ is trained on the frozen features $\phi(x)$ by minimizing the cross-entropy of $\mathrm{softmax}(W_p\,\phi(x)+b_p)$ over the $C$ classes, without updating the base model. The importance of neuron $n$ is derived from the probe weights, either globally or for a specific class $c$:
\begin{equation}I_{\mathrm{g}}(n) = \sum_{c=1}^{C} \lvert W_p[c,n]\rvert , \qquad I_{c}(n) = \lvert W_p[c,n]\rvert .\end{equation}
Given a selection percentage $p\in(0,1]$, the top-$k$ neurons by importance form the target set $S$, with $k$ defined in Eq.~\ref{eq1}. A global index $n$ is mapped to its layer and hidden dimension as $\ell = \lfloor n/H\rfloor$ and $d = n \bmod H$.

Interventions are applied during the forward pass through hooks acting on the selected coordinates $n\in S$. Let $a_n$ denote the activation of neuron $n$, $z_{c^\star}$ the output logit of a target class $c^\star$, and $W$ the classification-head weights. The operators are:
\begin{itemize}
\item \emph{Silencing}: $a_n \leftarrow 0$.
\item \emph{Gaussian noise}: $a_n \leftarrow a_n + \varepsilon_n$, with $\varepsilon_n\sim\mathcal{N}(0,\sigma^2)$.
\item \emph{Logit bias}: $z_{c^\star} \leftarrow z_{c^\star} + b$.
\item \emph{Weight edit}: $W[c^\star, S] \leftarrow W[c^\star, S] + \Delta$.
\item \emph{Bit-flip}: $w \leftarrow \mathrm{flipMSB}(w)$ for $w \in W[\cdot,S]$.
\end{itemize}
All operators are reversible: the original activations or parameters are restored after each run.

\subsection{Explainability Block}
The Explainability Block is the foundational component of SYNAPSE, designed to map the model's internal logic by quantifying the contribution of individual neurons to the classification task. This module performs a non-destructive extraction of internal activations (the numerical responses of neurons to specific inputs) to study how information is encoded across the Transformer layers without altering the model's parameters.

A critical technical element introduced at this stage is the \texttt{[CLS]} (Classification) token. In Transformer-based architectures, the \texttt{[CLS]} is a special token prepended to every input sequence. After passing through the encoder layers, the final hidden state is designed to serve as a summarized representation of the entire sequence for classification.

To validate the framework's versatility across different domains, two distinct datasets are employed. First, a tokenized version of the \textbf{MalwSpecSys} dataset is used, containing system call sequences labeled as \textit{Normal}, \textit{Bashlite}, \textit{Bdvl}, \textit{RansomwarePoC}, or \textit{TheTick}. Second, the \textbf{GoEmotions} corpus is evaluated using a \textit{monologg/bert-base-cased-goemotions-original} model, focused on a six-class slice: \textit{anger}, \textit{disgust}, \textit{fear}, \textit{joy}, \textit{sadness}, and \textit{surprise}.

The decision to extract activations exclusively from the \texttt{[CLS]} token is driven by computational efficiency. Due to the significant length of input sequences (particularly in system call traces, which can produce extremely long token sequences), extracting activations for every token would incur prohibitive computational costs and inefficient data management. By focusing on the \texttt{[CLS]} token at each encoder block, the framework achieves a drastic reduction in dimensionality and processing time. While this approach moves away from token-level granularity, the resulting summary preserves the global context needed for accurate, scalable neuron importance ranking, ensuring the analysis remains feasible even for large-scale models.

\paragraph{Activation Extractor}
This block extracts the internal representations associated with the \texttt{[CLS]} token across all hidden layers of a Transformer model, serving as the basis for subsequent neuron-level analysis. First, each tokenized input is run through the model to extract the set of hidden states of every layer, apart from the embedding layer, which represents an initial representation of information and has no valuable information, and the last classification layer included in the model, which was discarded because of its obvious importance, presenting one neuron per label. When task-specific information is extracted from the remaining layers, the hidden states are stacked, and the \texttt{[CLS]} token is isolated for extraction.

\paragraph{Activation Loader}
This module loads previously extracted neural activations from disk and prepares them for subsequent analysis. For each syscall sequence, it returns an activation matrix that can be used as input for probing classifiers and neuron-level analyses.

\paragraph{Linear Probe Training}
A linear probe, a simple linear classifier used to evaluate the information encoded in the representations of a pre-trained model, is trained to be applied on top of the frozen \texttt{[CLS]} representations and neither alters the original model nor is it a substitute for it. Instead, it verifies whether these representations encode features that can be separated by the linear classifier. Probe accuracy is a direct measure of how effectively the network has distilled input information into linearly separable patterns aligned with the target labels.

\subsection{Analysis Block}
After training the probing classifier, two custom functions were implemented to provide more control over neuron selection. The aim of these functions is to obtain:
\begin{itemize}
\item \textbf{A global ranking}, identifying most important neurons across all classes combined. This method computes a ranking of all neurons by the sum of their absolute weight contributions across classes, and selects the top-$k$ neurons by count, where $k$ is the number of neurons to be further silenced.
\item \textbf{A label-specific ranking}, highlighting neurons particularly important for predicting each individual label. This approach computes importance weights independently for each output label, enabling a targeted selection of neurons.
\end{itemize}

These rankings guide the selection of neurons to silence during the experiments and enable a systematic evaluation of model robustness under various perturbation strategies.

\paragraph{Top-K Neuron Selection}
To evaluate the impact of internal representations on model behavior, the framework must first identify which units are most functionally relevant.

\textbf{Neuron Importance Estimation.} Neuron importance is quantified using the weights of a lightweight linear probe trained on the \texttt{[CLS]} activations extracted from each encoder layer. The probe assigns a weight to each neuron for each output class, serving as a proxy for its contribution to the classification decision. Two types of importance rankings are derived from these weights:
\begin{itemize}
    \item \textit{Global ranking}: The importance of a neuron is defined as the sum of the absolute values of its weights across all classes, representing label-agnostic relevance.
    \item \textit{Class-conditional ranking}: The importance is defined by the absolute value of the weight assigned to a specific class $c$, representing label-specific relevance.
\end{itemize}

\textbf{Selection and Scope.} Based on a selection percentage $p \in (0,1]$, the framework identifies the top-$k$ neurons to be targeted. The value of $k$ depends on the chosen scope of the intervention~(Eq.~\ref{eq1}):
\begin{equation}\label{eq1}
k =
\begin{cases}
\lfloor p \cdot H \cdot L \rfloor & \text{for multi-layer scope} \\
\lfloor p \cdot H \rfloor & \text{for last-layer only scope}
\end{cases}
\end{equation}

where $H$ represents the hidden size and $L$ the number of layers. Global neuron indices are then mapped to their specific layer and local hidden dimension using integer division and modulo operations with respect to $H$.

\subsection{Adversarial Block} \label{subsec:attacks}
This section describes the suite of perturbations implemented to evaluate the model's internal resilience. The framework's attack surface comprises several strategies designed to probe how information loss affects classification: (i) Silencing, which mutes specific neurons to test functional redundancy; (ii) Noise Injection, which introduces stochastic perturbations; and (iii) Logit and Weight Manipulation, used for stress-testing under extreme conditions.

\textbf{Silencing policy}.
Silencing sets the activations of selected dimensions to zero in the encoder output during the forward pass. This process is non-destructive and training-free: no gradients are computed, and no model parameters are permanently modified. The primary objective of the silencing policy is to identify the degree of functional specialization within the model's architecture. By systematically muting neurons, one can determine whether specific labels depend on a small set of highly specialized units or whether task-relevant information is distributed across the network. Three silencing variants are implemented:
\begin{itemize}
    \item \textbf{Global Undirected Silencing:} Targets the top-$k$ neurons from the global ranking. The goal is to evaluate the model's overall stability after removing its most active units across all classes.
    \item \textbf{Global Directed Silencing:} Selects neurons that have the strongest influence on a specific target label according to the linear probe, but applies a zeroing intervention. This tests whether the model's ability to identify a particular class can be neutralized by silencing its most influential supporting neurons.
    \item \textbf{Per-Class Silencing:} Focuses exclusively on the top-$k$ neurons belonging to a specific class's ranking, probing class-conditional brittleness and potential leakage between labels.
\end{itemize}

\paragraph{Global Neuron Ablation Experiment (TH-3)}
The first type of experiment focused on the global importance of neurons. The top-ranked neurons across all classes were progressively silenced, starting with a small number of neurons and increasing incrementally. After each step, the model's overall classification performance was evaluated to assess how performance degradation correlates with silencing key neurons.\\

\paragraph{Label-Specific Neuron Ablation Experiment (TH-3)}
The second type of experiment is focused on label-specific importance. For each label, neurons identified as most critical to that label were progressively silenced. After each ablation step, the model's ability to correctly classify instances of that particular label was assessed, enabling a detailed analysis of the specialization and role of different neurons in the decision-making process.\\

\paragraph{Logit Bias (TH-4)}
This intervention perturbs the model \emph{outputs} directly by adding a constant offset to the logit of a target class just before the softmax. Concretely, a forward hook is installed on the model's output to add a bias $b$ to the logit of class $c^\star$ at inference time (optionally, a balanced variant subtracts a small amount from the remaining classes to increase contrast). No parameters are trained or permanently modified; the change is transient and removed after the run. This design emulates output-layer tampering used in white-box threat models while remaining architecture-agnostic and reversible. In this work, it is adapted to the neuron-centric setting by reporting how the induced shift interacts with the probe-driven class structure (e.g., the fraction of samples redirected to $c^\star$ and per-class F1-Score under attack).

\paragraph{Gaussian Noise (TH-3)}
This intervention injects zero-mean Gaussian noise into \texttt{[CLS]} activations of the encoder at selected neuron dimensions. Forward hooks are registered on the encoder's output blocks; at each block, the \texttt{[CLS]} vector is perturbed at the top-$k$ neuron indices provided by the probe ranking (global or class-conditional). The noise magnitude is governed by a standard deviation $\sigma$, and the scope can be last-layer only or all layers. Unlike classical baselines that add noise at the input embedding level, this variant places perturbations \emph{inside} the network and guides them by attribution, emphasizing neurons deemed influential by the probe. The attack is fully training-free and reversible, and it allows controlled sweeps over $(k,\sigma)$ to quantify robustness as a function of neuron importance.

\paragraph{Class-Directed Output Reweighting (TH-2)}
This intervention emulates weight-space tampering at inference time by temporarily editing the linear classification head. Let $W\in\mathbb{R}^{C\times H}$ and $b\in\mathbb{R}^{C}$ be the head's weights and bias (from hidden size $H$ to $C$ classes). A copy of $(W,b)$ is backed up; then, for a target class $c^\star$, small increments $\Delta$ are applied only on the \emph{columns} of $W$ that correspond to the top-$k$ neuron indices (global or class-conditional) mapped to hidden dimensions. Optionally, a balanced push reduces the same columns for non-target rows to enhance separation, and a selective suppression term can suppress a specific competitor class. A bias-only variant modifies $b$ instead of $W$. After inference, the original parameters are restored. This design adapts ideas from weight or bit-level fault attacks to a controlled, attribution-guided, training-free setting: modifications are confined to the head, targeted to probe-selected columns, and are strictly temporary, enabling clean ablation studies without retraining.

\paragraph{Attribution-Guided Bit-Flip (TH-3)}
This intervention emulates a transient hardware fault, such as a Rowhammer-induced memory disturbance~\cite{rowhammer}, by flipping individual bits of the model's parameters, in line with the bit-flip and hardware-fault attack literature~\cite{bitFlipAttack,terminalBrainDamage}. Instead of searching for the most damaging bits through gradient-guided optimization, the attack is guided by the probe-derived ranking: the top-$k$ most important weights, mapped from the neuron ranking to the columns of the classification head, are selected, and the most significant bit of their IEEE-754 exponent is flipped by reinterpreting each weight as an integer, applying an XOR mask, and casting it back. This produces the largest possible magnitude perturbation per flipped bit while keeping the intervention reversible and training-free, since the original parameters are restored after inference. Sweeping the fraction of flipped weights quantifies how many attribution-selected bit flips are required before the classifier collapses, yielding a fault-robustness budget rather than an optimal adversarial search.

\paragraph{Random Noise Injection (TH-1)}
As a baseline comparison, a simple random noise attack was implemented by directly adding noise to the model's input embeddings. This method is not gradient-based and does not rely on the model's internal computations. Instead, it introduces randomly generated perturbations to the input in an uninformed manner. This technique is commonly used in the literature as a baseline for evaluating model robustness \cite{zhao2018generating,xu2018feature}.
The intensity of the noise is controlled by a parameter epsilon ($\epsilon$), which scales the amplitude of the random values added to each embedding vector. Higher values of $\epsilon$ lead to greater distortions and typically more severe performance degradation. In this work, noise was applied directly to the input embeddings during inference, without modifying the model's parameters or input token sequence. For each input sequence, input embeddings are computed and perturbed by additive Gaussian noise scaled by a user-defined factor $\epsilon$. When Gaussian noise is injected into the embeddings, each token's position in the original vector is shifted, subtly disrupting semantic relationships and the overall similarity structure among embeddings. As a result, the following layers receive distorted representations, leading to less coherent internal activations and, ultimately, reduced classification performance.\\

\paragraph{Fast Gradient Sign Method (TH-1)}
FGSM is a single-step adversarial attack proposed by Goodfellow et al. \cite{goodfellow2015explainingharnessingadversarialexamples}. It works by adding a perturbation to the input in the direction of the loss function's gradient, thereby maximizing the loss. This perturbation is scaled by a small factor $\epsilon$ that controls its magnitude. The lower the value of $\epsilon$, the less perceptible the perturbation is. Unlike the random noise injection approach proposed in Subsection~\ref{subsec:attacks}, which perturbs the input without considering the model's behavior, FGSM generates targeted perturbations based on the model's internal gradients, making it more effective and model-aware. To apply this approach to the scenario studied in this work, an FGSM variant was implemented and directly applied to the input embeddings of tokenized syscall traces. First, the cross-entropy loss is backpropagated to compute gradients with respect to each embedding vector. Then, a perturbation is created by taking the element-wise sign of that gradient and scaling it by a predefined $\epsilon$ to maximize loss. Finally, this signed perturbation is added to the original embeddings, yielding adversarial embeddings, which are then passed through the model to assess the resulting degradation in classification performance. The model is then tested with input examples to confirm the degradation, using state-of-the-art evaluation metrics.

\section{Validation} \label{sec:val}

A rigorous and reproducible evaluation protocol was established to ensure fair and meaningful comparisons across models, datasets, and experimental settings. The protocol standardizes data preprocessing, model configuration, and evaluation metrics, providing a consistent basis for analyzing the effects of neuron-level interventions under controlled conditions.

\subsection{Experimental setup}

\paragraph{Setting}
The experiments reported in this work were executed on an NVIDIA DGX Spark node, inside the vendor-provided NVIDIA PyTorch container (CUDA, Python~3.12), which supplies the GPU-enabled build of PyTorch; the Transformers library is pinned to a fixed commit so that the software environment is reproducible. The framework carries no GPU-specific requirement and was developed on a consumer machine (Apple M4 Pro, 24~GB of RAM), on which every intervention also runs; the GPU node was used to repeat the evaluations over multiple seeds within a practical time budget. The wall-clock and peak-memory figures reported in Section~\ref{sec:results} correspond to the DGX Spark node. The corresponding code for every described procedure is publicly available in~\cite{kikaymusic2025repo}.

\paragraph{Scenario}
This work evaluates inference-time, white-box interventions at the neuron level (hidden representation dimensions) using PyTorch forward hooks. In this setting, an analyst has read-write access to intermediate activations during the forward pass but neither re-trains the model nor alters the training data.

\paragraph{Metric}
The F1-Score was chosen as the primary evaluation metric because it balances precision and recall into a single value, ensuring that both false positives and false negatives are properly accounted for. More precisely, the macro-averaged F1-Score was used. By doing so, every label is given equal weight regardless of its support, ensuring that performance impact is measured consistently across all classes and preventing majority-class dominance from obscuring degradation in less frequent categories.

\paragraph{Perturbation magnitudes}
The intervention magnitudes are not single, arbitrary choices. The silencing fraction $p$ is swept over a wide range, from 5\% to 95\% of the ranked neurons, and the input-noise level $\epsilon$ is swept over $\{0.1, 0.5, 1.0\}$, so the values reported in the tables are points on these sweeps rather than isolated selections. The remaining single-value operating points are deliberate stress-test settings. The Gaussian-noise configuration ($\sigma=0.9$ applied to 60\% of the most salient neurons) is a high-variance, majority-ablation stress point chosen to probe robustness under strong but unbiased internal perturbations. The logit-bias and weight-space increments were set to the largest values that still produce a measurable and interpretable redirection of predictions before the model collapses into a degenerate all-target regime, a behavior observed empirically; they are therefore reported as controlled operating points rather than tuned hyperparameters.

\subsection{Experimental datasets}

To evaluate the proposed approach across heterogeneous domains, experiments are conducted in two classification scenarios: malware detection from system-call sequences and emotion classification from natural-language text. These settings are selected to cover structurally diverse input modalities and tasks while relying on established, reproducible evaluation pipelines.

\paragraph{Malware detection}
The malware experiments are based on the token-based detection pipeline introduced by Sánchez et al.~\cite{sanchez2024transfer}, which models system-call traces as token sequences processed by a Transformer encoder followed by a linear classification head. This pipeline is adopted because it constitutes a well-established and validated baseline for sequence-based malware detection, enabling comparison with prior work while avoiding confounding architectural modifications.

The dataset consists of system-call sequences labeled as \textit{Bashlite}, \textit{Bdvl}, \textit{Normal}, \textit{RansomwarePoC}, and \textit{TheTick}. The encoder architectures considered include \textit{BERT}, \textit{BigBird}, \textit{DistilBERT}, and \textit{Longformer}, covering both standard and long-sequence Transformer variants. The original model described in that work includes an additional final classification layer with one neuron per target class; this layer is excluded from the analysis, as directly manipulating output units would trivially affect predictions and would not provide insight into the internal representations learned by the encoder.

\paragraph{Emotion classification (GoEmotions)}
To complement the malware detection setting with a natural language task, experiments are also conducted on the GoEmotions dataset introduced by Demszky et al.~\cite{demszky2020goemotions}, using the publicly available \textit{monologg/bert-base-cased-goemotions-original} model. The GoEmotions dataset provides annotations for 28 emotion categories; to obtain a controlled and interpretable setting, a six-class subset consisting of \textit{anger}, \textit{disgust}, \textit{fear}, \textit{joy}, \textit{sadness}, and \textit{surprise} is constructed. Only single-label samples are retained, and the original 28-dimensional output logits are restricted to the selected classes, with predictions obtained by taking the argmax over this subset.

For controlled experimental sweeps, a stratified evaluation set balanced across classes is employed, with selected configurations replicated on larger batches when computational resources permit. Tokenization follows the original model tokenizer, using standard padding and truncation strategies.

\subsection{Model representation choice}

The selection of BERT-based architectures relies on the use of the \texttt{[CLS]} token, which serves as a condensed representation of the entire sequence's contextual information. By focusing on the activations of the \texttt{[CLS]} token, the proposed framework significantly improves computational efficiency by avoiding the overhead of extracting data for each individual token while preserving the global features required for accurate neuron-level interpretability and classification analysis. Table~\ref{tab:model_specs} summarizes the architectural specifications of the evaluated models. The sufficiency of this summary representation is not assumed: Section~\ref{sec:results} contrasts the \texttt{[CLS]}-based neuron ranking with the one obtained from mean-pooled token representations, which aggregate information from every position in the sequence, and reports their agreement in Table~\ref{tab:cls-meanpool}.

To probe architectural generality beyond the \texttt{[CLS]}-based encoders, the evaluation also includes GPT-2, a decoder-only model reused from the original malware detection pipeline. Because decoder-only architectures do not expose a \texttt{[CLS]} token, GPT-2 is summarized with the mean-pooled representation defined in Section~\ref{sec:fw}. Its native context is limited to 1{,}024 tokens, so the long system-call traces are truncated to their first 1{,}024 tokens, in contrast to the 4{,}096-token window of BigBird and Longformer. This reduced context, together with the averaging induced by mean pooling, is later shown to affect the reliability of the neuron ranking for this model, while leaving its overall detection performance competitive.

Accordingly, GPT-2 is evaluated on the subset of interventions that decides whether the protocol transfers to a decoder-only architecture: activation extraction and probe-based ranking under mean pooling, the random-neuron control, the agreement with gradient-based attribution, the global silencing sweep, and the bit-flip attack. The input- and weight-space perturbations act
on the input embeddings or on the classification head and are therefore independent of how the sequence is summarized, so replicating them on GPT-2 would add cost without testing the architectural claim; they are reported for the encoder models only.
%%%

\begin{table}[!t]
\centering
\caption{Architectural specifications of the evaluated Transformer models.}
\renewcommand{\arraystretch}{1.3}
\resizebox{\columnwidth}{!}{%
\begin{tabular}{lccccc}
\hline
\textbf{Model} & \makecell{\textbf{Layers} \\ \textbf{($L$)}} & \makecell{\textbf{Hidden} \\ \textbf{Size ($d_{model}$)}} & \makecell{\textbf{Total Neurons} \\ \textbf{(MLP)}} & \makecell{\textbf{Max. Seq.} \\ \textbf{Length}} & \makecell{\textbf{Param.} \\ \textbf{Count}} \\
\hline
BERT (base)       & 12 & 768 & 36,864 & 512   & 110M \\
BigBird           & 12 & 768 & 36,864 & 4,096 & 128M \\
DistilBERT        & 6  & 768 & 18,432 & 512   & 66M  \\
Longformer        & 12 & 768 & 36,864 & 4,096 & 149M \\
GPT-2 (decoder)  & 12 & 768 & 36,864 & 1,024 & 124M \\
GoEmotions (BERT) & 12 & 768 & 36,864 & 512   & 110M \\
\hline
\end{tabular}}
\label{tab:model_specs}
\end{table}

\section{Results} \label{sec:results}
This section presents a comprehensive evaluation of the experimental results obtained using the SYNAPSE framework. The analysis is structured into three main pillars: (i) Baseline Performance, establishing a reference for each architecture; (ii) Neuron-level Interventions, transitioning from global performance degradation to targeted class-specific silencing; and (iii) Comparative Stress-tests, where neuron-centric attacks are benchmarked against traditional gradient-based methods, noise injection, and weight-space manipulations. This progression enables characterization of how neuron importance, intervention scope, and domain-specific sensitivities influence the robustness of Transformer-based classifiers.

\subsection{Malware Detection}
This section presents the experimental results obtained with the proposed framework in the malware detection scenario, covering the different evaluation settings and intervention strategies considered in this study.

Before conducting any experiments, the baseline performance of the original models was evaluated using the MalwSpecSys dataset. As shown in Table~\ref{tab:tab1Orig}, models designed to handle longer sequences, such as Longformer and BigBird, achieve the highest macro-F1 scores (0.875 and 0.879, respectively). In contrast, BERT and DistilBERT perform worse (0.601 and 0.576), likely due to the fixed context window limitation when processing complex system call traces, while the decoder-only GPT-2 lies in between (0.625). Under the normal-vs-malicious detection view, all evaluated models nonetheless reach a ROC-AUC of at least 0.83, confirming detection performance well above chance. These values provide a clean reference point for quantifying the degradation caused by subsequent silencing strategies.

\subsubsection{Global Silencing}
\label{sec:global_sil}
The first stage of the analysis considers Global Silencing, a class-agnostic approach that progressively disables a fixed proportion of the most salient neurons across the entire model. The objective is to assess the overall structural redundancy of the network: if performance drops sharply with minimal silencing, the model relies on a group of smart neurons, whereas if the drop is gradual, the knowledge is considered widely distributed.

\begin{figure}[!t]
    \centering
    \includegraphics[width=\linewidth]{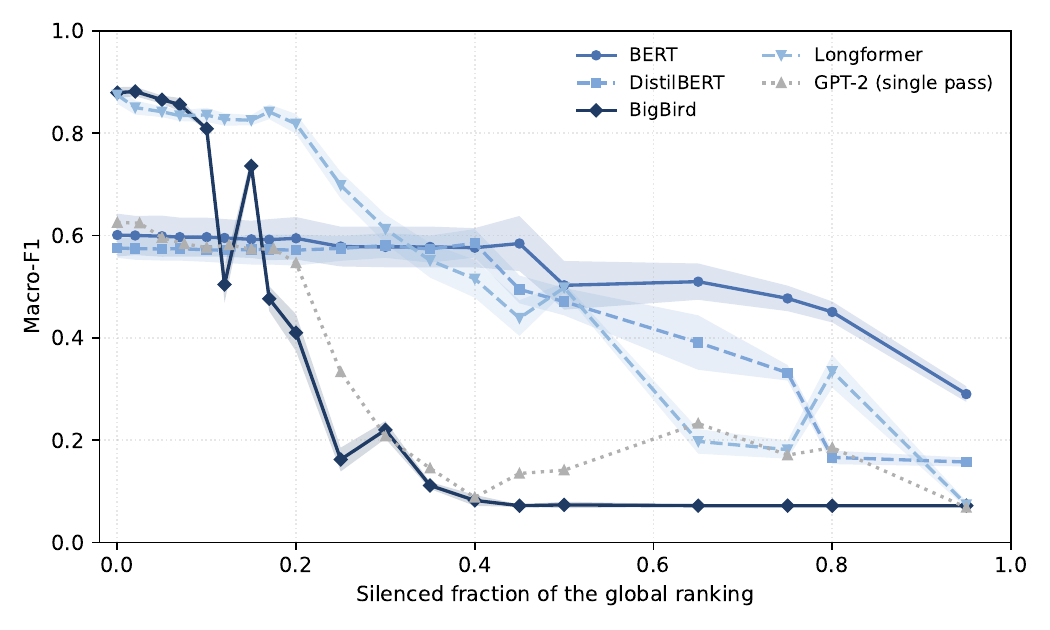}
    \caption{Macro-F1 under global neuron silencing for every evaluated model. Shaded bands span one standard deviation over the five seeds; GPT-2 was measured in a single pass and is therefore shown without a band.}
    \label{fig:silencing-global}
\end{figure}

As seen in Figure~\ref{fig:silencing-global}, the hooks successfully intercepted the forward pass, leading to a noticeable and predictable degradation, and the narrowness of the seed bands shows that the shape of each curve, not merely its endpoint, is stable across repetitions. For BigBird, performance stays close to its baseline while up to roughly 10\% of the most salient neurons are silenced, but it then collapses sharply: the macro-F1 score falls from $0.809{\pm}0.003$ at 10\% to $0.410{\pm}0.036$ at 20\% and to $0.074{\pm}0.006$ at 50\%, well below a random-guess level. The descent through that band is not smooth: performance drops to $0.504{\pm}0.036$ at 12\%, recovers to $0.736{\pm}0.017$ at 15\%, and falls again to $0.476{\pm}0.024$ at 17\%. The seed dispersion at those points is far too small for the oscillation to be sampling noise, so it is a reproducible property of the ranking rather than an artefact: removing a slightly larger set of neurons can leave the classifier in a better state than removing a smaller one, which indicates that some units in the mid-ranking region act as compensators for those removed before them. The framework surfaces this because it sweeps the ranking rather than testing a single budget, and it is a concrete caution for any defence that assumes degradation to be monotone in the number of ablated units. Beyond that point the model degenerates into predicting a single surviving class (TheTick), which keeps the macro-F1 from reaching zero while the remaining classes drop to an F1 close to zero. This indicates that, for long-context architectures, task-relevant information is concentrated in a compact set of intermediate neurons whose removal cannot be compensated by the untouched final classification layer, in contrast to the more gradual decline exhibited by BERT and DistilBERT, which retain roughly half of their baseline macro-F1 even at a 95\% budget. Table~\ref{tab:global-silencing} provides a detailed comparison of this global attack across all tested architectures.

\begin{table*}[!t]
    \centering
    \caption{Global neuron silencing: macro-F1-Score for different Transformer models under increasing fractions of globally important neurons being silenced. $\Delta$F1-Score denotes the relative variation with respect to the baseline, at a 95\% silenced fraction. Encoder values are the mean over five seeds on the same balanced subset as Table~\ref{tab:tab1Orig}; the standard deviation never exceeds $0.054$ at any point of the sweep, and the per-condition dispersion for the reported fractions is given in Table~\ref{tab:tab1Orig}. GPT-2 was measured in a single pass.}
    \label{tab:global-silencing}
    \small
    \setlength{\tabcolsep}{4pt} % <-- Reducimos el espacio entre columnas
    \begin{tabularx}{\textwidth}{@{} l >{\hsize=1.4\hsize\centering\arraybackslash}X *{7}{>{\hsize=0.94\hsize\centering\arraybackslash}X} r @{}}
        \toprule
        \textbf{Model} & \textbf{Baseline} & \textbf{5\%} & \textbf{10\%} & \textbf{20\%} & \textbf{30\%} & \textbf{50\%} & \textbf{75\%} & \textbf{95\%} & \textbf{$\Delta$F1 (\%)} \\
        \midrule
        BERT       & 0.601 & 0.599 & 0.597 & 0.594 & 0.578 & 0.503 & 0.477 & 0.290 & $-51.7$ \\
        BigBird    & 0.879 & 0.865 & 0.809 & 0.410 & 0.220 & 0.074 & 0.072 & 0.072 & $-91.8$ \\
        DistilBERT & 0.576 & 0.574 & 0.572 & 0.571 & 0.580 & 0.471 & 0.332 & 0.157 & $-72.7$ \\
        Longformer & 0.875 & 0.842 & 0.835 & 0.818 & 0.613 & 0.498 & 0.182 & 0.074 & $-91.5$ \\
        GPT-2 & 0.625 & 0.596 & 0.576 & 0.546 & 0.208 & 0.141 & 0.171 & 0.069 & $-89.0$ \\
        \bottomrule
    \end{tabularx}
\end{table*}

\subsubsection{Detection View of Global Silencing}
Macro-F1 measures how well the five classes are separated, but it does not say what an operator would experience were a detector of this kind placed in service. Table~\ref{tab:tab1Orig} therefore reports, alongside the baseline of each model, the same normal-versus-malicious view under silencing at three fractions that bracket the transition: 10\% (before the collapse), 20\% (within it), and 50\% (well past it). Reporting both conditions in a single table keeps the comparison anchored to the undisturbed operating point of each model, and every row is computed on the same balanced evaluation subset and averaged over the same five seeds as the sweep of Table~\ref{tab:global-silencing}. Here FAR is the fraction of benign traffic flagged as malicious and MAR the fraction of malicious traffic missed as normal, both read at the arg-max decision, while ROC-AUC, EER, and TPR@1\%FPR are derived from the receiver operating characteristic of the malicious score (one minus the normal-class probability).

The detection view surfaces a property of the audited models that macro-F1 alone conceals, and it illustrates the kind of diagnostic the framework is designed to produce. The degradation is not a graceful loss of accuracy but a saturation of the decision: for BigBird, FAR climbs from 0.019 at a 10\% budget to 0.957 at 20\% and to 1.000 at 50\%, while MAR falls to zero. The classifier does not become uncertain, it becomes degenerate, assigning every input to a single malware class. MAR near zero should not be read as a benefit here, since a classifier that alarms unconditionally misses nothing by construction and carries no information; the collapse of TPR@1\%FPR from 0.826 to 0.329 over the same range confirms that no usable operating point survives. The per-class breakdown locates the effect precisely: at a 50\% budget every class except one falls to zero recall, and the surviving class retains a precision of only 0.32, which identifies the majority class of the evaluation subset as the attractor of the degenerate decision rather than any learned preference. Longformer degrades along the same path but later, and the 512-token encoders, whose macro-F1 was lower to begin with, hold a comparatively stable operating point up to a 50\% budget. GPT-2 follows the long-context pattern, with TPR@1\%FPR vanishing to zero at a 50\% budget.

Read as an audit result, this is actionable rather than merely descriptive. It tells a practitioner that the higher-capacity detectors concentrate their decision in a subset of neurons whose removal does not blur the boundary but voids it, and that the failure would announce itself as a flood of false alarms rather than as silent evasion. Both the affected subset and the fraction at which the transition occurs are reported by the framework, thereby making them available as targets for a subsequent hardening or retraining stage.

These metrics are reported for the global silencing sweep, the intervention through which the framework establishes its main claim, and for the baseline of every model. They are deliberately not extended to the per-family breakdown or to the emotion task: neither is a detection problem, and a false-alarm rate has no operational meaning where there is no alarm to raise. Accuracy and F1-Score remain the appropriate measures in those settings.

\begin{table*}[!t]
\centering
\scriptsize
\caption{Detection view of the malware models, at the baseline and under increasing fractions of globally silenced neurons, collapsing the five classes into a normal-vs-malicious view. Macro-F1 remains the primary multiclass metric. Encoder values are the mean $\pm$ standard deviation over five seeds; GPT-2 was measured in a single pass.}
\label{tab:tab1Orig}

\renewcommand{\arraystretch}{1.05}
\setlength{\tabcolsep}{2pt}

\begin{tabularx}{\textwidth}{@{}l c c c c c c c@{}}
\toprule
\textbf{Model} & \textbf{Silenced} & \textbf{Macro-F1} & \textbf{ROC-AUC} & \textbf{EER} & \textbf{FAR} & \textbf{MAR} & \textbf{TPR@1\%} \\
\midrule
BERT       & Baseline & $0.601{\pm}0.042$ & $0.830{\pm}0.017$ & $0.280{\pm}0.024$ & 0.011 & 0.375 & 0.632 \\
           & 10\% & $0.597{\pm}0.038$ & $0.833{\pm}0.015$ & $0.272{\pm}0.012$ & 0.005 & 0.382 & 0.629 \\
           & 20\% & $0.594{\pm}0.042$ & $0.849{\pm}0.019$ & $0.252{\pm}0.008$ & 0.005 & 0.391 & 0.623 \\
           & 50\% & $0.503{\pm}0.048$ & $0.831{\pm}0.024$ & $0.262{\pm}0.017$ & 0.005 & 0.406 & 0.598 \\
\midrule
DistilBERT & Baseline & $0.576{\pm}0.019$ & $0.851{\pm}0.018$ & $0.241{\pm}0.018$ & 0.020 & 0.379 & 0.642 \\
           & 10\% & $0.572{\pm}0.023$ & $0.847{\pm}0.022$ & $0.261{\pm}0.034$ & 0.025 & 0.380 & 0.640 \\
           & 20\% & $0.571{\pm}0.030$ & $0.844{\pm}0.024$ & $0.263{\pm}0.036$ & 0.035 & 0.380 & 0.635 \\
           & 50\% & $0.471{\pm}0.027$ & $0.823{\pm}0.028$ & $0.278{\pm}0.034$ & 0.380 & 0.239 & 0.615 \\
\midrule
BigBird    & Baseline & $0.879{\pm}0.011$ & $0.976{\pm}0.002$ & $0.080{\pm}0.014$ & 0.070 & 0.089 & 0.879 \\
           & 10\% & $0.809{\pm}0.003$ & $0.970{\pm}0.004$ & $0.091{\pm}0.016$ & 0.019 & 0.159 & 0.826 \\
           & 20\% & $0.410{\pm}0.036$ & $0.644{\pm}0.048$ & $0.386{\pm}0.048$ & 0.957 & 0.053 & 0.242 \\
           & 50\% & $0.074{\pm}0.006$ & $0.778{\pm}0.049$ & $0.268{\pm}0.054$ & 1.000 & 0.000 & 0.329 \\
\midrule
Longformer & Baseline & $0.875{\pm}0.017$ & $0.973{\pm}0.012$ & $0.087{\pm}0.024$ & 0.089 & 0.081 & 0.865 \\
           & 10\% & $0.835{\pm}0.014$ & $0.969{\pm}0.014$ & $0.094{\pm}0.028$ & 0.045 & 0.132 & 0.856 \\
           & 20\% & $0.818{\pm}0.019$ & $0.958{\pm}0.015$ & $0.110{\pm}0.029$ & 0.046 & 0.134 & 0.798 \\
           & 50\% & $0.498{\pm}0.009$ & $0.704{\pm}0.027$ & $0.348{\pm}0.038$ & 0.728 & 0.207 & 0.484 \\
\midrule
GPT-2 & Baseline & 0.625 & 0.885 & 0.181 & 0.119 & 0.323 & 0.614 \\
           & 10\% & 0.576 & 0.868 & 0.215 & 0.000 & 0.380 & 0.652 \\
           & 20\% & 0.546 & 0.825 & 0.282 & 0.024 & 0.399 & 0.468 \\
           & 50\% & 0.141 & 0.368 & 0.620 & 0.976 & 0.051 & 0.000 \\
   &  \\
\bottomrule
\end{tabularx}
\end{table*}

\subsubsection{Random-Neuron Control}
To verify that the probe-derived ranking identifies functionally relevant neurons rather than arbitrary ones, a control experiment silences, at each fraction, a set of randomly selected neurons of the same size $k$ and compares the resulting degradation with that of the probe-ranked top-$k$ neurons. Table~\ref{tab:random-control} reports both conditions at a silenced fraction of 50\% over five seeds, together with the empirical $p$-value and the $z$-score of the top-$k$ degradation relative to the random distribution. For the four encoder models, silencing the top-$k$ neurons degrades the macro-F1 far more than silencing $k$ random neurons; for BigBird, for instance, performance falls to 0.073 against 0.781 under random selection ($z=14.2$, $p=0.048$), confirming that the ranking captures units that are disproportionately important for the task. GPT-2 is the exception: under its mean-pooled representation, random silencing is at least as damaging as the probe-guided selection ($z<0$, not significant), which indicates that the ranking is less reliable for the decoder-only model, consistent with the weaker agreement between its probe ranking and gradient-based attribution measures.

\begin{table}[!t]
\centering
\caption{Random-neuron control at a silenced fraction of 50\% (five seeds). Top-$k$ F1 is the macro-F1 after silencing the probe-ranked neurons; Random-$k$ F1 is the mean $\pm$ std after silencing the same number of randomly chosen neurons. The $z$-score and empirical $p$-value quantify how far the top-$k$ degradation lies from the random distribution.}
\label{tab:random-control}
\setlength{\tabcolsep}{6pt}
\begin{tabularx}{\columnwidth}{@{}X c c c c@{}}
\toprule
\textbf{Model} & \textbf{Top-$k$ F1} & \textbf{Random-$k$ F1} & \textbf{$z$-score} & \textbf{$p$} \\
\midrule
BERT       & 0.437 & $0.543{\pm}0.006$ & 17.5 & 0.048 \\
DistilBERT & 0.465 & $0.571{\pm}0.005$ & 22.5 & 0.048 \\
BigBird    & 0.073 & $0.781{\pm}0.050$ & 14.2 & 0.048 \\
Longformer & 0.485 & $0.874{\pm}0.019$ & 20.8 & 0.048 \\
GPT-2      & 0.141 & $0.085{\pm}0.043$ & $-1.3$ & 0.833 \\
\bottomrule
\end{tabularx}
\end{table}

\subsubsection{Attribution-Method Agreement}
As a further check on the meaningfulness of the probe-based ranking, and following the suggestion to compare against alternative attribution techniques, the probe ranking is contrasted with two neuron-level attribution methods: conductance, the integrated-gradients-based measure of neuron importance\addtxt{ of Dhamdhere et al.~\cite{dhamdhere2018conductance}}, and activation-times-gradient\addtxt{, in the formulation of Shrikumar et al.~\cite{shrikumar2018efficient}; the linear-probe ranking that serves as the reference follows the layer-wise probing methodology of NeuroX~\cite{dalvi2023neurox}, on which the framework of this work is built}. Standard integrated gradients and SHAP are feature-attribution methods, so their neuron-level counterparts are used here. The agreement on the most important neurons is measured through Spearman rank correlation ($\rho$) and top-$k$ overlap at $k=10\%$, and reported in Table~\ref{tab:attribution}. For the encoder models the probe ranking correlates positively and substantially with both gradient-based measures ($\rho$ between 0.44 and 0.71, highest for Longformer), indicating that the lightweight probe recovers largely the same salient neurons as considerably more expensive attribution methods. Consistent with the random-neuron control and the truncated mean-pooled representation discussed above, GPT-2 is again the exception, with only a weak correlation ($\rho\approx0.14$). The probe ranking is retained as the operational method throughout the interventions because it is the one designed for the sequence-level \texttt{[CLS]} representation, is lightweight, and is architecture-agnostic across the evaluated encoders; causal mediation analysis is left as a complementary direction for future work. The last row of Table~\ref{tab:attribution} reports a further agreement check of the same nature, contrasting the \texttt{[CLS]} ranking with the one obtained from mean-pooled token representations; it is discussed in Section~\ref{sec:results} together with the emotion-classification results, where that comparison was carried out.

\begin{table*}[!t]
\centering
\small
\caption{Agreement between the probe-based neuron ranking and alternative references, over $9{,}216$ neurons. The first two blocks contrast the probe with two neuron-level attribution methods on the malware task; the last row contrasts the \texttt{[CLS]}-based ranking with the one derived from mean-pooled token representations on GoEmotions. Rank correlations are computed on the full ranking; overlap and Jaccard index on the top-$k$ neurons at $k=10\%$ ($k=922$), for which chance-level overlap is $0.10$.}
\label{tab:attribution}
\label{tab:cls-meanpool}% alias: la referencia del texto original marcado con \change sigue resolviendo
\setlength{\tabcolsep}{5pt}
\begin{tabularx}{\textwidth}{@{}Xcccc@{}}
\toprule
\textbf{Comparison} & $\boldsymbol{\rho}$ & $\boldsymbol{\tau}$ & \textbf{Overlap@10\%} & \textbf{Jaccard} \\
\midrule
\multicolumn{5}{@{}l}{\textit{Probe ranking vs.\ conductance (malware)}} \\
\quad BERT       & 0.515 & 0.349 & 0.225 & 0.126 \\
\quad DistilBERT & 0.629 & 0.437 & 0.213 & 0.119 \\
\quad BigBird    & 0.464 & 0.321 & 0.325 & 0.194 \\
\quad Longformer & 0.701 & 0.509 & 0.394 & 0.245 \\
\quad GPT-2      & 0.146 & 0.097 & 0.141 & 0.076 \\
\midrule
\multicolumn{5}{@{}l}{\textit{Probe ranking vs.\ activation\,$\times$\,gradient (malware)}} \\
\quad BERT       & 0.437 & 0.294 & 0.177 & 0.097 \\
\quad DistilBERT & 0.623 & 0.430 & 0.239 & 0.135 \\
\quad BigBird    & 0.495 & 0.348 & 0.670 & 0.504 \\
\quad Longformer & 0.710 & 0.520 & 0.498 & 0.331 \\
\quad GPT-2      & 0.137 & 0.091 & 0.110 & 0.058 \\
\midrule
\multicolumn{5}{@{}l}{\textit{\texttt{[CLS]} vs.\ mean-pooled representation (GoEmotions)}} \\
\quad BERT (GoEmotions) & 0.419 & 0.285 & 0.325 & 0.194 \\
\bottomrule
\end{tabularx}
\end{table*}

\addtxt{\subsubsection{Intervention Effectiveness of Alternative Neuron Rankings}}

\addtxt{Table~\ref{tab:attribution} quantifies which neurons each criterion selects, but agreement between rankings does not determine what each of them achieves once it drives an intervention. That question is addressed through a controlled comparison using the global silencing protocol described in Section~\ref{sec:global_sil}, with the neuron ordering supplied by four different criteria. Within this comparison, the evaluation sample, the five seeds, the swept suppression fractions, the forward-hook mechanism, and the macro-F1 metric are common to all four arms. Since the neuron ordering is the only variable, the comparison isolates its contribution. The criteria span three families. The first is the probe-weight ranking used throughout this work, which follows the probing methodology of NeuroX~\cite{dalvi2023neurox}. The second is Probeless (Antverg and Belinkov ~\cite{antverg2022pitfalls}), a ranking from the same library that orders neurons by the separation of their class-conditional mean activations and therefore requires no probe. The remaining two are the gradient-based measures of Table~\ref{tab:attribution}, the conductance of Dhamdhere et al.~\cite{dhamdhere2018conductance}, and the activation-times-gradient measure of Shrikumar et al.~\cite{shrikumar2018efficient}. Effectiveness is measured relative to the probe arm using paired comparisons within the same seed and evaluation sample. This design reduces sampling variability and isolates the effect of neuron ordering. Table~\ref{tab:ranking-effectiveness} reports the additional degradation each criterion produces over the probe, averaged across the swept fractions.}

\addtxt{All three alternatives cause more damage than the probe, and the margin is substantial: on BERT, ordering neurons by activation-times-gradient costs the model a further $0.33$ macro-F1 on average across the sweep. No criterion is uniformly best. Activation-times-gradient leads on BERT, DistilBERT, and Longformer, whereas on BigBird conductance is the most damaging by a clear margin, and on Longformer the three alternatives fall within a narrow band of one another. The ordering criterion therefore matters as much as the intervention that follows it, and which criterion is preferable depends on the model. The probe ranks below the other three methods because it measures a different quantity. It ranks neurons by how well class information can be linearly decoded from their activations, rather than by how much they influence the model’s output. Silencing tests the latter directly, while gradient-based measures approximate it. The probe is nonetheless retained as the operational ranking for the reasons stated in Section~\ref{sec:fw}, namely that it is defined on the sequence-level representation, and requires no gradients: its training takes under one second on every model (Table~\ref{tab:complexity}), whereas conductance requires a forward and a backward pass for every layer and integration step of each example, amounting to two hundred and forty such passes per example at the settings used here. This comparison shows that the framework does not depend on that choice: the protocol admits any per-neuron ordering and measures its consequences under a common experimental condition, allowing criteria from different families to be placed on the same scale.}

\begin{table}[!t]
\centering
\caption{\addtxt{Effectiveness of alternative neuron-ranking criteria under global silencing,
relative to the probe ranking. Each entry is the additional macro-F1 degradation the criterion
produces over the probe across the eighteen swept fractions, averaged over five seeds; positive
values indicate a criterion that concentrates more damage than the probe. Each arm is compared
against the probe arm of the same seed and the same evaluation sample, so that only the neuron
ordering differs.}}
\label{tab:ranking-effectiveness}
\setlength{\tabcolsep}{6pt}
\small
\addtxt{\begin{tabularx}{\columnwidth}{@{}Xcc@{}}
\toprule
\textbf{Ranking} & \textbf{BERT} & \textbf{DistilBERT} \\
\midrule
Probe (this work)            & \multicolumn{2}{c}{\textit{reference}} \\
Activation $\times$ gradient & $+0.334{\pm}0.021$ & $+0.240{\pm}0.011$ \\
Conductance                  & $+0.146{\pm}0.020$ & $+0.114{\pm}0.009$ \\
Probeless                    & $+0.086{\pm}0.010$ & $+0.047{\pm}0.009$ \\
\midrule
\textbf{Ranking} & \textbf{BigBird} & \textbf{Longformer} \\
\midrule
Probe (this work)            & \multicolumn{2}{c}{\textit{reference}} \\
Activation $\times$ gradient & $+0.209{\pm}0.022$ & $+0.208{\pm}0.040$ \\
Conductance                  & $+0.302{\pm}0.021$ & $+0.141{\pm}0.040$ \\
Probeless                    & $+0.130{\pm}0.020$ & $+0.175{\pm}0.045$ \\
\bottomrule
\end{tabularx}}
\end{table}

\subsubsection{Per-Class Silencing}

\begin{figure*}[!t]
    \centering
    \begin{subfigure}[t]{0.32\textwidth}
        \centering
        \includegraphics[width=\linewidth]{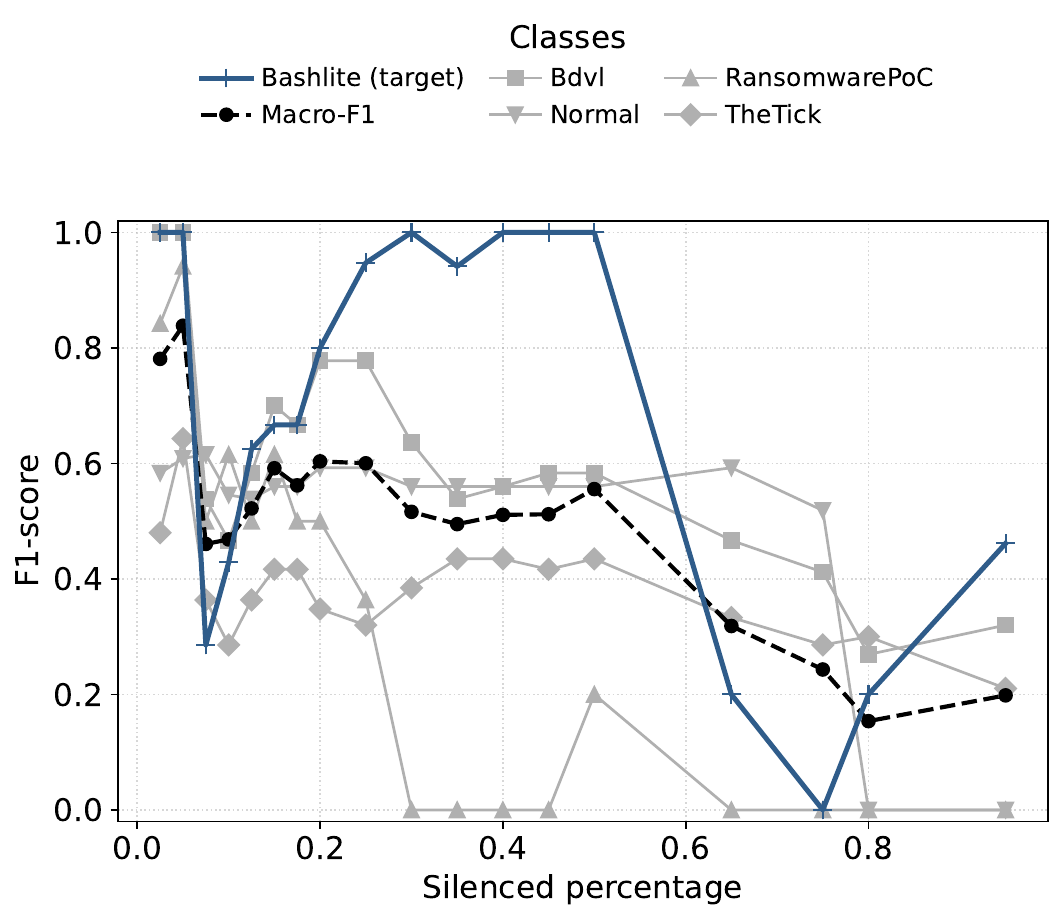}
        \caption{Class 0 (Bashlite)}
        \label{fig:class0}
    \end{subfigure}\hfill
    \begin{subfigure}[t]{0.32\textwidth}
        \centering
        \includegraphics[width=\linewidth]{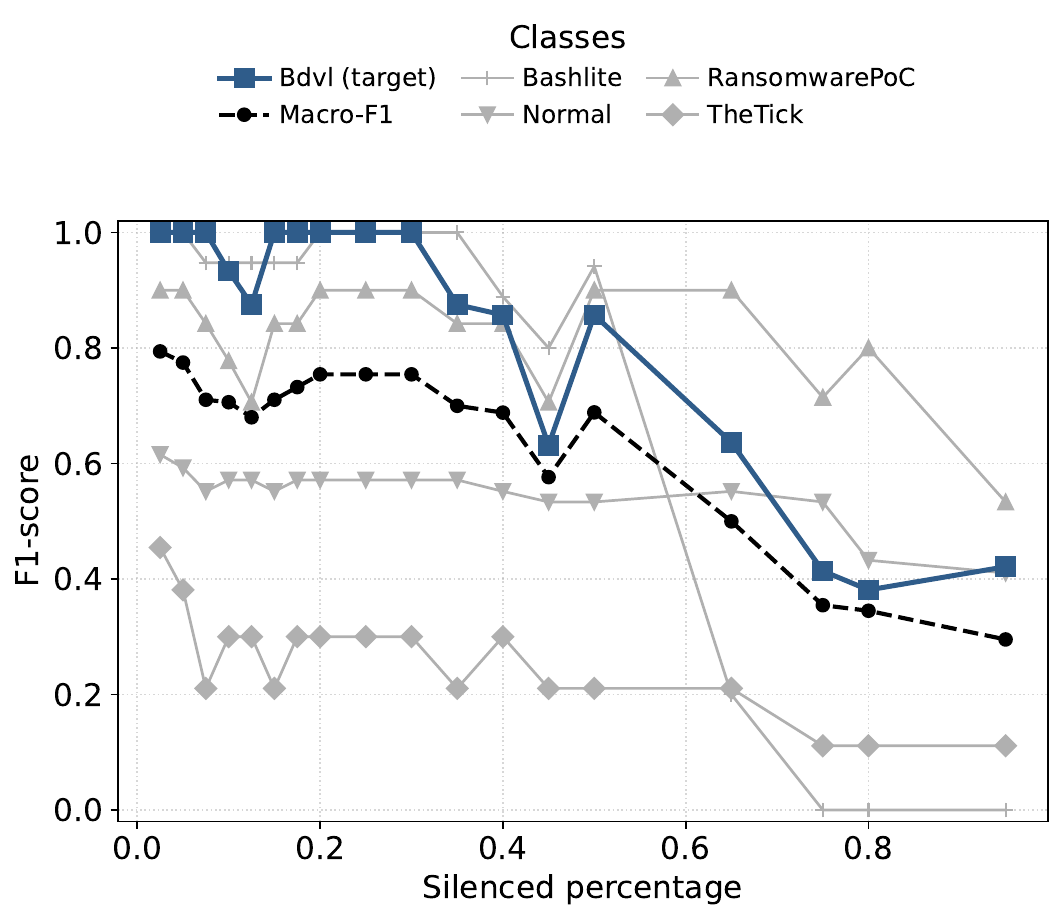}
        \caption{Class 1 (Bdvl)}
        \label{fig:class1}
    \end{subfigure}\hfill
    \begin{subfigure}[t]{0.32\textwidth}
        \centering
        \includegraphics[width=\linewidth]{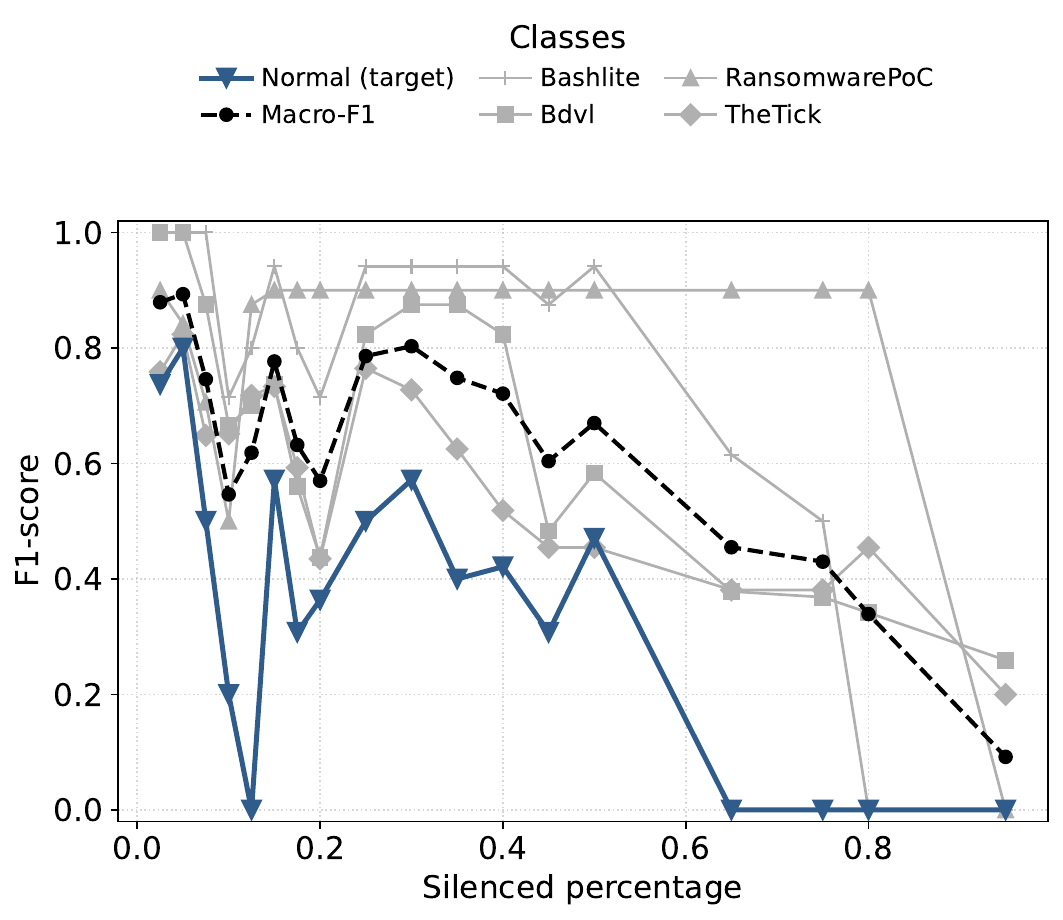}
        \caption{Class 2 (Normal)}
        \label{fig:class2}
    \end{subfigure}

    \vspace{0.6em}

    \begin{subfigure}[t]{0.32\textwidth}
        \centering
        \includegraphics[width=\linewidth]{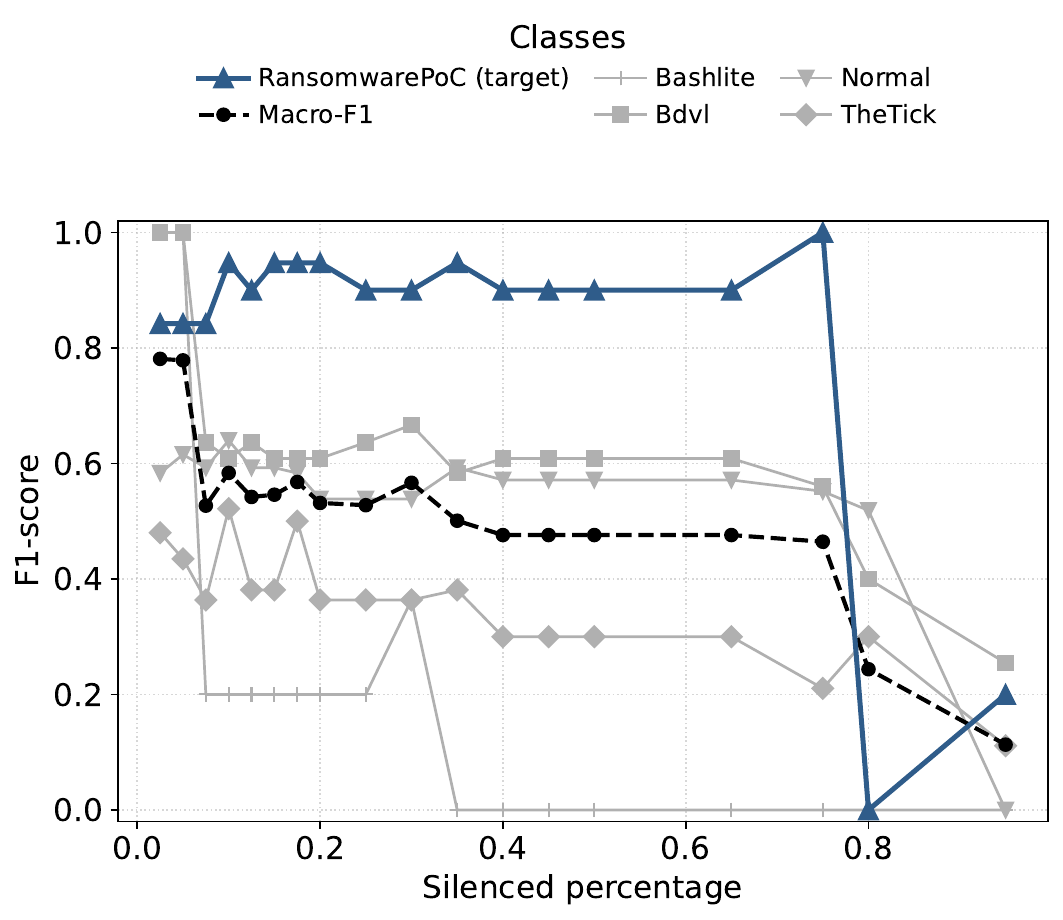}
        \caption{Class 3 (RansomwarePoC)}
        \label{fig:class3}
    \end{subfigure}
    \hspace{0.03\textwidth}
    \begin{subfigure}[t]{0.32\textwidth}
        \centering
        \includegraphics[width=\linewidth]{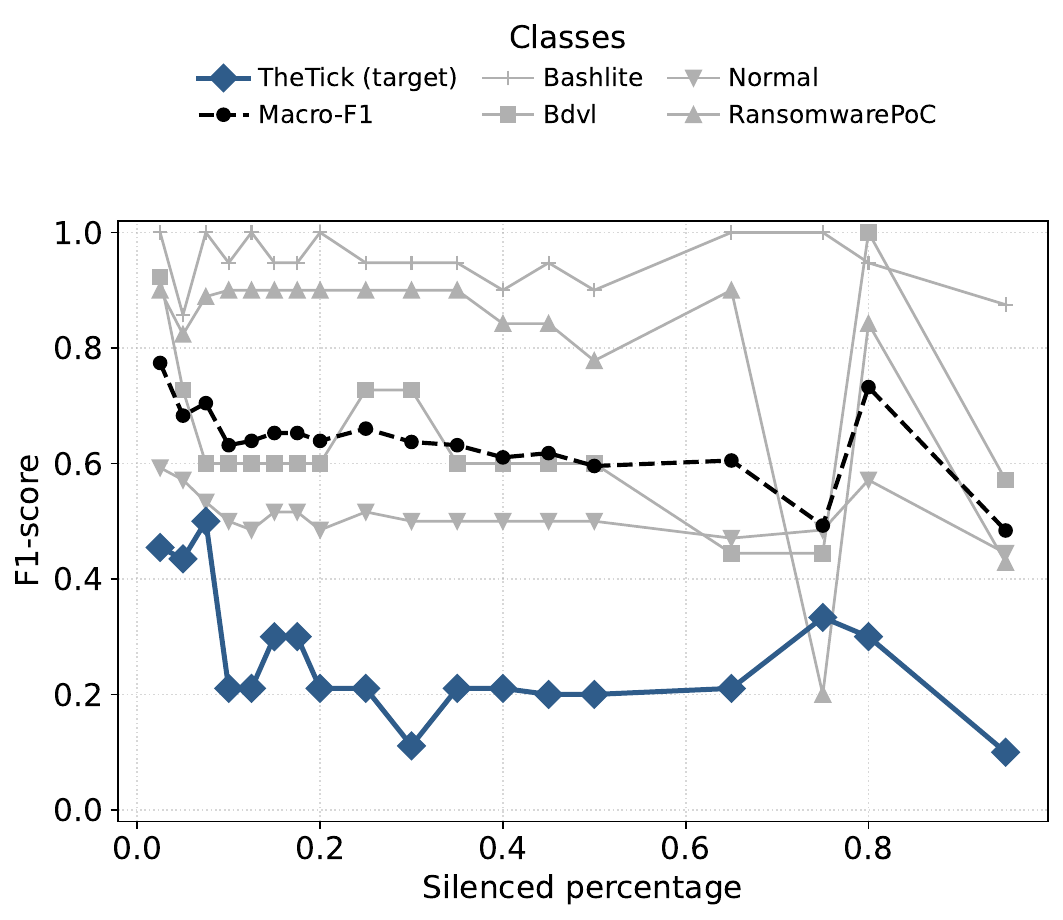}
        \caption{Class 4 (TheTick)}
        \label{fig:class4}
    \end{subfigure}

    \caption{Results after silencing per-class ranking for BigBird across classes.}
    \label{fig:bigbird_perclass_silencing}
\end{figure*}

Figures~\ref{fig:class0}-\ref{fig:class4} show the F1-Score degradation (per target label) as progressively more salient neurons are silenced for each syscall traffic class, and Table~\ref{tab:class_silencing_final} summarizes the end point of every sweep, reporting for each model and each target class the F1-Score of the targeted class and the weighted-overall F1 at a 95\% silencing budget. Across most targets, the curves drop substantially, most sharply for \textit{Normal} and \textit{TheTick} (Figures~\ref{fig:class2} and \ref{fig:class4}), indicating that the model depends on relatively small subsets of neurons. In addition, while the F1-score for the targeted label decreases, other labels often degrade as well, and overall performance decreases (Figures~\ref{fig:class0}-\ref{fig:class4}), consistent with the distributed nature of the representations.

Looking at the raw activations extracted in the first section of the pipeline, the highest activations were around 0.6--0.7, and neurons showed positive activations for up to three different labels. This supports the idea that no tiny, fully isolated neuron groups can be identified for a more precise attack, and it also explains why other labels are affected when targeting a specific one (Figures~\ref{fig:class0}-\ref{fig:class4}).

A particularly relevant case is \textit{TheTick} (Figure~\ref{fig:class4}): silencing the neurons ranked as most important for this label drives its F1-Score down to about 0.1, while most other traffic classes remain comparatively high. This indicates that the components driving decisions on \textit{TheTick} are relatively localized, so this label can be neutralized while keeping other predictions largely intact.

The benign \textit{Normal} class (Figure~\ref{fig:class2}) is instead the most fragile: once its top neurons are silenced its F1-Score collapses to 0, which also depresses the weighted-overall performance, showing that the neurons supporting normal traffic are entangled with the rest of the decision structure. By contrast, \textit{RansomwarePoC} (Figure~\ref{fig:class3}) is comparatively robust against neuron-level attacks, since degrading it requires silencing a large fraction of neurons and it stays relatively stable when other labels are targeted.

\begin{table*}[t]
    \centering
\footnotesize
    \caption{Final performance under class-targeted neuron silencing: F1-Score of the targeted class (F1) and weighted-overall F1 (WO), at a 95\% silencing budget of the class-specific ranking. Classes are listed by their (alphabetical) label name.}
    \label{tab:class_silencing_final}
    \begin{tabularx}{\textwidth}{@{} l *{10}{>{\centering\arraybackslash}X} @{}}
        \toprule
        \textbf{Model} & \multicolumn{2}{c}{\textbf{Bashlite}} & \multicolumn{2}{c}{\textbf{Bdvl}} & \multicolumn{2}{c}{\textbf{Normal}} & \multicolumn{2}{c}{\textbf{Rans.PoC}} & \multicolumn{2}{c}{\textbf{TheTick}} \\
        \cmidrule(lr){2-3} \cmidrule(lr){4-5} \cmidrule(lr){6-7} \cmidrule(lr){8-9} \cmidrule(l){10-11}
        & \textbf{F1} & \textbf{WO*} & \textbf{F1} & \textbf{WO} & \textbf{F1} & \textbf{WO} & \textbf{F1} & \textbf{WO} & \textbf{F1} & \textbf{WO} \\
        \midrule
        BERT       & 0.571 & 0.513 & 0.923 & 0.533 & 0.000 & 0.470 & 0.800 & 0.473 & 0.605 & 0.442 \\
        BigBird    & 0.462 & 0.200 & 0.421 & 0.258 & 0.000 & 0.104 & 0.200 & 0.109 & 0.100 & 0.420 \\
        DistilBERT & 0.409 & 0.312 & 0.857 & 0.337 & 0.250 & 0.387 & 0.200 & 0.370 & 0.552 & 0.370 \\
        Longformer & 0.000 & 0.314 & 0.389 & 0.287 & 0.125 & 0.500 & 0.200 & 0.273 & 0.286 & 0.425 \\
        \bottomrule
        \multicolumn{11}{l}{\footnotesize *WO: Weighted Overall.}
    \end{tabularx}
\end{table*}

Across all Transformer models, class-targeted silencing produces highly asymmetric degradation patterns, indicating that each model exhibits localized, unevenly distributed class-specific sensitivities. Three consistent trends emerge. First, models differ sharply in their reliance on class-specialized neurons: BERT and DistilBERT retain a high F1-Score for \textit{Bdvl} even when its most influential neurons are suppressed (0.923 and 0.857, respectively), whereas the benign \textit{Normal} class collapses to 0.000 in both BERT and BigBird, indicating a concentrated representation in which a small set of neurons carries most of its discriminative power. Second, degradation does not correlate with global model strength: despite its strong baseline, BigBird suffers the most severe class-wise collapses (\textit{Normal} at 0.000 and \textit{TheTick} at 0.100), suggesting that its high overall accuracy relies on highly specialized neurons that, once removed, produce brittle behavior, and the weighted-overall column confirms that suppressing one class also drags down the others. Third, each model exhibits a distinct profile of robust and fragile classes: BERT is resilient for \textit{Bdvl} and \textit{RansomwarePoC} (0.923 and 0.800) but fragile for \textit{Normal} (0.000), whereas Longformer is the only model whose \textit{Bashlite} class collapses entirely (0.000) while degrading more gradually elsewhere. Across architectures, \textit{Bdvl} is consistently the most robust label and \textit{Normal} the most fragile.

Overall, these results indicate that class-specific neuron dependence is 
highly model-dependent and unevenly distributed, supporting the hypothesis 
that Transformer classifiers do not allocate representational capacity 
uniformly across labels. Instead, each architecture develops idiosyncratic 
islands of specialization, whose ablation exposes how the internal class 
structure emerges from a sparse subset of critical neurons.

\subsubsection{Classical Perturbation Baselines}
In addition to neuron-level interventions, gradient-based attacks provide a complementary perspective on model vulnerability. In this context, the Fast Gradient Sign Method (FGSM) was adapted to operate under the constraints of the proposed framework, enabling a direct comparison between neuron-centric perturbations and gradient-driven attacks. Alongside it, two unstructured baselines are considered: random noise injected at the embedding level, swept over its magnitude $\epsilon$, and Gaussian noise injected into the \texttt{[CLS]} representation of the most influential neurons. The three share the same logic, a perturbation of growing magnitude applied without any target class, and are therefore reported together in Table~\ref{tab:classical_perturbations}. They provide the reference against which the cost of the neuron-level and weight-level interventions should be read.

\begin{table*}[!t]
\centering
\small
\caption{Classical perturbation baselines: weighted F1-Score under FGSM, random noise injection at the embedding level, and Gaussian noise injected into the \texttt{[CLS]} representation. FGSM is reported at the single budget $\epsilon=0.1$ standard for this attack; the 512-token models (BERT, DistilBERT, Longformer) share the same 50-sample evaluation subset, while BigBird uses its own 4096-token subset. All values are single-pass measurements on the 50-sample evaluation subset, following the protocol of the original experiments, and are therefore not directly comparable with the macro-F1 of Tables~\ref{tab:tab1Orig} and~\ref{tab:global-silencing}.}
\label{tab:classical_perturbations}
\label{tab:fgsm_comparison}% alias
\label{tab:random_noise}% alias
\label{tab:gaussian-noise}% alias
\renewcommand{\arraystretch}{1.15}
\setlength{\tabcolsep}{6pt}
\begin{tabularx}{\textwidth}{@{}l l cccc@{}}
\toprule
\textbf{Perturbation} & \textbf{Level} & \textbf{BERT} & \textbf{DistilBERT} & \textbf{BigBird} & \textbf{Longformer} \\
\midrule
FGSM  & $\epsilon=0.1$  & 0.370 & 0.352 & 0.395 & 0.353 \\
\midrule
Random noise & $\epsilon=0.1$ & 0.6834 & 0.6081 & 0.8056 & 0.8712 \\
                      & $\epsilon=0.5$ & 0.6860 & 0.6081 & 0.7438 & 0.3903 \\
                      & $\epsilon=1.0$ & 0.6135 & 0.5214 & 0.0344 & 0.0344 \\
             & $\Delta$F1 (\%) & $-10.2$ & $-14.3$ & $-95.9$ & $-96.0$ \\
\midrule
Gaussian noise & $\sigma=0.9$, 60\% & 0.7438 & 0.5812 & 0.7510 & 0.7443 \\
\bottomrule
\end{tabularx}
\end{table*}

FGSM is reported at a single perturbation budget, $\epsilon=0.1$, the value most commonly adopted for this attack in the literature. At this operating point all four models fall to a weighted F1-Score between $0.35$ and $0.40$, a substantial loss of class separation relative to their clean baselines. Gradient-based perturbation injected at the input-embedding level is therefore uniformly effective across architectures: unlike the neuron- and weight-level interventions, which separate the models by capacity and context length, FGSM at this budget leaves no architecture markedly more robust than the others.

\begin{figure}[!t]
    \centering
    \includegraphics[width=\linewidth]{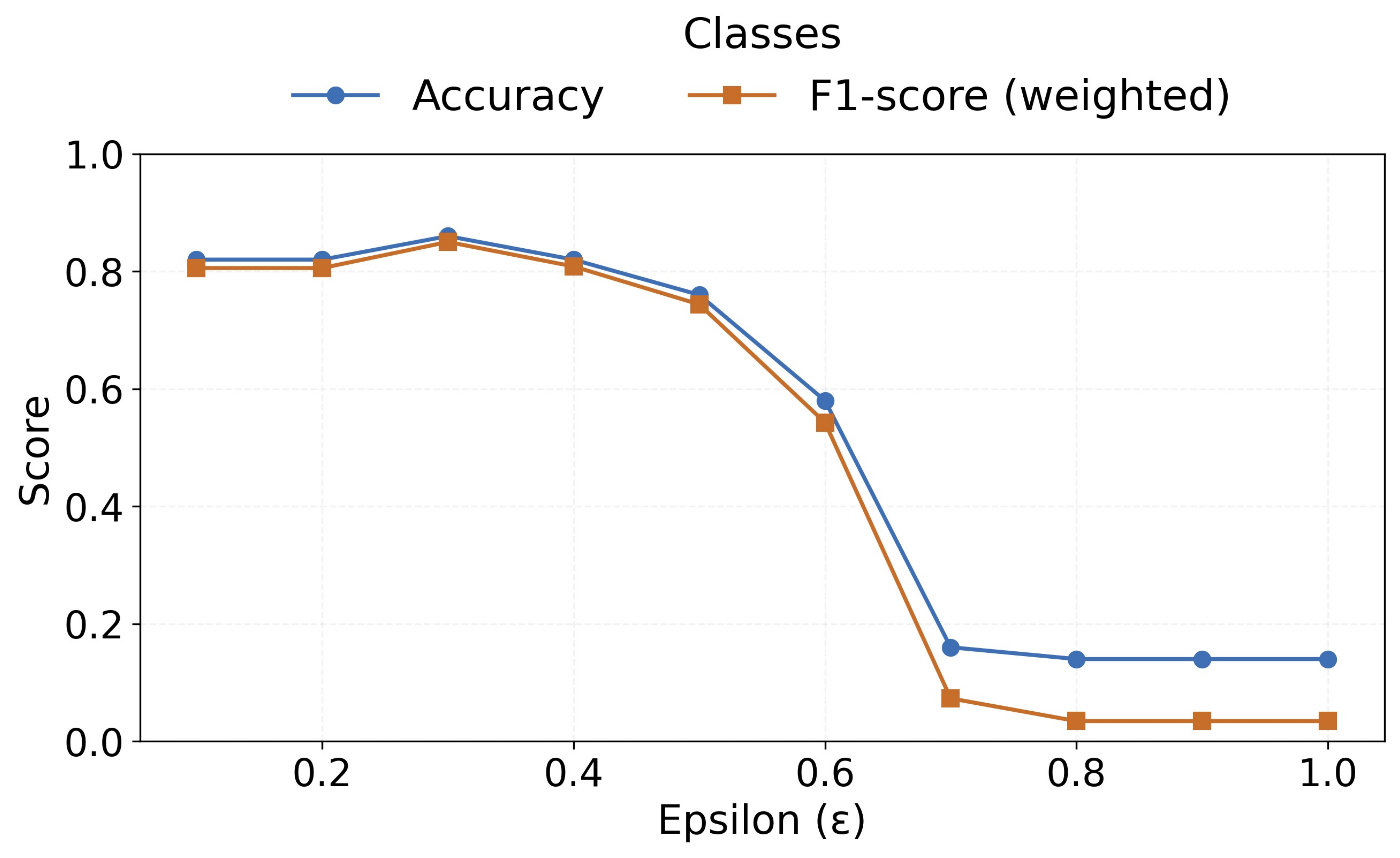}
    \caption{Random Noise model performance degradation}
    \label{fig:random_noise}
\end{figure}

Random noise injection exposes markedly different robustness profiles across the four Transformer models.
BigBird serves as an illustrative case through the sweep shown in Figure~\ref{fig:random_noise}.
For small perturbations ($\epsilon \leq 0.4$), both accuracy and weighted F1-Score remain essentially unchanged, oscillating around the baseline and indicating that BigBird is able to absorb low-magnitude unstructured perturbations at the embedding level. Once the perturbation exceeds a critical threshold ($\epsilon \approx 0.6$), performance collapses sharply: both metrics drop by more than half, and for higher noise levels ($\epsilon \geq 0.8$) the model enters a near-failure regime. This behavior suggests that random perturbations disrupt the geometry of the embedding space only after a tipping point is reached, beyond which the internal attention structure can no longer maintain separability between classes.

The comparative results across models show that this transition separates the two long-context architectures from the two 512-token ones. BigBird and Longformer both absorb low-magnitude noise and then collapse almost entirely once the tipping point is passed, converging to an F1-Score of 0.0344 at $\epsilon=1.0$ (drops of $-95.9\%$ and $-96.0\%$), with Longformer breaking earlier, around $\epsilon=0.5$, than BigBird, which holds until $\epsilon\approx0.6$. BERT and DistilBERT, in contrast, degrade only mildly across the whole sweep ($-10.2\%$ and $-14.3\%$ at $\epsilon=1.0$), never leaving the neighbourhood of their baselines. The ordering therefore mirrors the one observed under neuron silencing and under the bit-flip attack: the higher-capacity, long-context models achieve the stronger baselines but concentrate their decision in a more brittle configuration, whereas the compact encoders trade accuracy for stability. That three interventions acting on entirely different surfaces, the input embeddings, the intermediate activations, and the weights of the head, converge on the same ranking is the strongest evidence in this study that the effect is a property of the audited models rather than of any single probing technique.

The impact of Gaussian perturbations on the \texttt{[CLS]} representation is relatively mild across all models.
Even with a high noise level ($\sigma = 0.9$) applied to $60\%$ of the most influential neurons, the weighted F1-Scores remain close to their baselines, with no model experiencing severe degradation. At the same time, BigBird and Longformer maintain the highest robustness (F1-Score $\approx 0.75$ and $0.74$, respectively), and DistilBERT shows the largest drop but still retains more than half of its predictive performance.

This pattern suggests that the models' internal representations redundantly distribute semantic information across multiple neural dimensions, making them resilient to high-variance perturbations as long as the perturbations are zero-mean and unbiased. Unlike targeted manipulations (e.g., logit bias or weight-space attacks), Gaussian perturbations do not systematically push predictions toward specific classes, which explains the relatively stable behavior observed for all architectures.

\subsubsection{Output- and Weight-Space Attacks}
The interventions considered so far act on internal activations. This subsection groups the four attacks that instead manipulate the final decision stage, either by shifting the logits directly (TH-4) or by editing the weights of the classification head (TH-2). They share a common experimental structure, a single configuration per model and a target class $c^\star=3$, and are therefore reported together in Table~\ref{tab:weight_output_attacks}: the logit-bias attack, the balanced weight push, the class-suppression variant, and the bias-only variant. Because these interventions aim to redirect predictions rather than to destroy performance, the informative quantities are not only the loss in F1-Score but also the fraction of non-target samples flipped to $c^\star$ and the overall share of predictions ending in $c^\star$.

\begin{figure}[!t]
    \centering
    \includegraphics[width=0.75\linewidth]{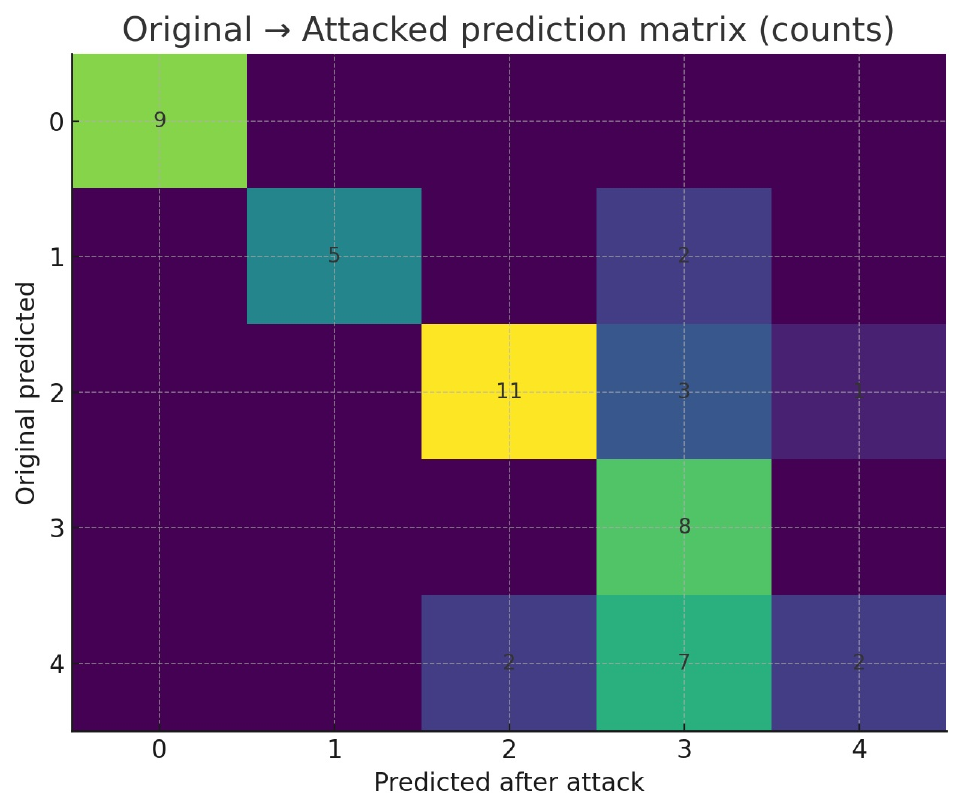}
    \caption{Logit Bias Attack results on BigBird}
    \label{fig:logitbiasmatrix}
\end{figure}

\begin{table*}[!t]
\centering
\scriptsize
\caption{Output- and weight-space attacks with target class $c^\star=3$, consolidating the logit-bias intervention (TH-4) and the three classification-head edits (TH-2). Flips$\rightarrow$Target is the fraction of non-target samples redirected to $c^\star$; Pred$\rightarrow$Target is the overall share of predictions assigned to $c^\star$. All values are single-pass measurements on the 50-sample evaluation subset and use the weighted F1-Score, following the protocol of the original experiments.}
\label{tab:weight_output_attacks}
\label{tab:logit_bias_models}% alias: mantiene resoluble la referencia del texto original marcado con \change
\renewcommand{\arraystretch}{1.15}
\setlength{\tabcolsep}{2pt}
\begin{tabularx}{\textwidth}{@{}l l *{4}{>{\centering\arraybackslash}X}@{}}
\toprule
\textbf{Attack} & \textbf{Model} & \textbf{Accuracy} & \textbf{F1-Score (weighted)} & \textbf{Flips$\rightarrow$Target (\%)} & \textbf{Pred$\rightarrow$Target (\%)} \\
\midrule
Logit bias & BERT       & 0.700 & 0.687 & 28.57 & 40 \\
($b=8.0$)  & BigBird    & 0.720 & 0.705 & 28.13 & 36 \\
                    & DistilBERT & 0.680 & 0.590 & 6.82  & 18 \\
                    & Longformer & 0.640 & 0.634 & 42.86 & 52 \\
\midrule
Balanced weight push & BERT       & 0.760 & 0.683 & 2.27  & 14 \\
($\Delta=0.3$)       & BigBird    & 0.380 & 0.313 & 71.43 & 76 \\
                              & DistilBERT & 0.560 & 0.415 & 18.18 & 28 \\
                              & Longformer & 0.820 & 0.807 & 0.00  & 6  \\
\midrule
Class suppression & BERT       & 0.760 & 0.683 & 2.27  & 14 \\
($\Delta=0.2$)    & BigBird    & 0.760 & 0.674 & 14.29 & 28 \\
                           & DistilBERT & 0.680 & 0.587 & 4.55  & 16 \\
                           & Longformer & 0.840 & 0.832 & 0.00  & 6  \\
\midrule
Bias-only & BERT       & 0.760 & 0.683 & 2.27 & 14 \\
($\Delta=0.6$) & BigBird    & 0.820 & 0.806 & 7.14 & 22 \\
                   & DistilBERT\textsuperscript{$\dagger$} & -- & -- & -- & -- \\
                   & Longformer & 0.860 & 0.852 & 7.14 & 22 \\
\bottomrule
\end{tabularx}
\footnotesize $^{\dagger}$ DistilBERT is omitted because the probe assigned zero importance to all neurons, preventing the bias-only attack from being instantiated.
\end{table*}

The logit-bias intervention produces a characteristic redistribution of predictions toward $c^\star$ while preserving the overall structure of the decision boundaries. Figure~\ref{fig:logitbiasmatrix} shows the original$\rightarrow$attacked prediction mapping for BigBird under a bias of $b=8$. The diagonal remains partially intact, indicating that a subset of samples is robust to perturbations in the output layer. However, several off-diagonal blocks become populated, most notably the transitions $2\!\rightarrow\!3$ and $4\!\rightarrow\!3$, which reflect a systematic drift of multiple classes toward the target label. This behavior is consistent with an additive shift applied directly to the output logits: classes with logits close to $c^\star$ are the first to flip. In contrast, well-separated classes remain stable unless the bias is sufficiently large. Across models, Table~\ref{tab:weight_output_attacks} shows that all architectures redirect a non-trivial fraction of predictions to the target class, but the intensity of this redirection varies markedly. DistilBERT is the most resistant, with only $18\%$ of predictions assigned to the target and a low flip rate of $6.82\%$. In contrast, Longformer shows the strongest drift, with $52\%$ of predictions redirected to the target and $42.86\%$ of non-target samples flipping, suggesting a more compressible logit geometry. BERT and BigBird lie in an intermediate regime, exhibiting substantial but not catastrophic reweighting of outputs.

The three weight-space variants reverse this ordering, which is the most informative contrast in the table. This variant produces strong directional shifts in BigBird, which becomes highly susceptible to target-class redirection (76\% of all predictions collapse to $c^\star$, with 71.4\% flips from non-target classes). In contrast, Longformer remains almost unaffected under the balanced push. By explicitly weakening a single competitor class, the class-suppression variant yields more controlled behavior: BigBird exhibits reduced collapse relative to the balanced push, while Longformer remains stable. On the other hand, DistilBERT responds to suppression with a moderate drift. The bias-only variant produces the mildest effects: models whose logits already provide clear inter-class separation, such as Longformer and BigBird, show minimal performance loss, confirming that a shift of the bias vector alone cannot reshape a decision geometry that the weight matrix keeps well separated. BERT stands apart from the other models: the balanced push, the class-suppression variant, and the bias-only edit leave its predictions identical to one another and to the clean baseline (weighted F1-Score $0.683$, $2.27\%$ non-target flips, $14\%$ of predictions on $c^\star$), indicating that at the tested edit scale its classification head absorbs the perturbation entirely rather than exhibiting a graded response.

Overall, these results demonstrate that weight-space perturbations expose architecture-dependent vulnerabilities: BigBird is consistently the most vulnerable model, Longformer the most resistant, DistilBERT occupies an intermediate regime, and BERT proves refractory to all three head edits, its predictions remaining fixed at the clean baseline. The contrast among variants also shows that attacks that modify the structure of $W$ are far more damaging than those that affect only the biases, confirming that the head's geometric configuration is a primary driver of robustness. Read against the logit-bias rows, the two families are complementary: an attacker able to shift logits reaches every model to some degree, whereas one editing the head reaches BigBird almost completely and Longformer barely at all. An audit that reported only one of the two would therefore misrank the architectures.

\subsubsection{Attribution-Guided Bit-Flip}
Table~\ref{tab:bitflip} reports the macro-F1 after flipping the most significant exponent bit of the top-$k$ most important weights, for increasing fractions of flipped parameters. The attack is markedly more efficient than neuron silencing: flipping as little as 2\% of the attribution-selected weights already collapses the long-context models, with BigBird and Longformer falling from about 0.88 to 0.16--0.18, and by a 5\% budget BERT, BigBird, Longformer, and GPT-2 all reach a near-degenerate regime around 0.07--0.12. DistilBERT is the most resilient, tolerating a 2\% budget almost unchanged (0.578) and collapsing only beyond a 10\% budget. This confirms that a handful of well-placed bit flips, selected purely from the probe ranking, are sufficient to destroy the classifier, echoing the graceless degradation reported for hardware fault attacks~\cite{terminalBrainDamage} and the supply-chain tampering threat model~\cite{badnets}. The vulnerability is specific to the malware detectors rather than a universal property of Transformers: under the same attack, the GoEmotions model is essentially unaffected, its F1-Score dropping by at most a few percent even at a 20\% budget, which suggests that the compactness and class structure of the malware task make its decision boundary far more fragile to bit-level corruption.

\begin{table*}[!t]
\centering
\footnotesize
\caption{Attribution-guided bit-flip attack: macro-F1 after flipping the most significant exponent bit of the top-$k$ most important weights, for increasing fractions of flipped weights. Values for the four encoders are the mean $\pm$ standard deviation over five seeds; GPT-2 was measured in a single pass and therefore carries no dispersion estimate. $\Delta$F1 is the relative variation at a 20\% budget. GPT-2 uses a first step of 2.5\% instead of 2\%.}
\label{tab:bitflip}
\setlength{\tabcolsep}{2pt}
\begin{tabularx}{\textwidth}{@{}X c c c c c c@{}}
\toprule
\textbf{Model} & \textbf{Baseline} & \textbf{2\%} & \textbf{5\%} & \textbf{10\%} & \textbf{20\%} & \textbf{$\Delta$F1 (\%)} \\
\midrule
BERT       & $0.601{\pm}0.042$ & $0.197{\pm}0.021$ & $0.092{\pm}0.015$ & $0.070{\pm}0.019$ & $0.071{\pm}0.014$ & $-88.2$ \\
DistilBERT & $0.576{\pm}0.019$ & $0.578{\pm}0.022$ & $0.371{\pm}0.010$ & $0.175{\pm}0.034$ & $0.077{\pm}0.021$ & $-86.7$ \\
BigBird    & $0.879{\pm}0.011$ & $0.157{\pm}0.012$ & $0.067{\pm}0.013$ & $0.065{\pm}0.011$ & $0.065{\pm}0.011$ & $-92.6$ \\
Longformer & $0.875{\pm}0.017$ & $0.179{\pm}0.019$ & $0.067{\pm}0.013$ & $0.065{\pm}0.011$ & $0.065{\pm}0.011$ & $-92.6$ \\
GPT-2 & 0.625 & 0.217 & 0.121 & 0.072 & 0.072 & $-88.5$ \\
\bottomrule
\end{tabularx}
\end{table*}

\subsection{Emotion Detection}

To complement the malware-oriented experiments, the framework is also evaluated on a fundamentally different domain: affective text classification. For this, this work employs the public monologg/bert-base-cased-goemotions-original model, trained on Google's GoEmotions corpus \cite{demszky2020goemotions}. As described in the experimental setup, the original 28-emotion space is reduced to a six-class configuration (anger, disgust, fear, joy, sadness, surprise). Only single-label examples are retained, and the model's 28-dimensional output is remapped by restricting the logits to the six relevant indices and selecting the class with the highest score within this subset. A stratified and balanced mini-evaluation set is used to ensure stable estimation across perturbation sweeps.

While the malware experiments analyze each attack type in dedicated subsections, the GoEmotions case follows a different reporting strategy. Since the six-class model is not the focus of the interpretability study, which is a cross-domain stress test, the objective is to assess how the attacks generalize beyond system-call classifiers. For this reason, all neuron-level and weight-level perturbations are consolidated into a single summary table. This unified presentation simplifies cross-attack comparison, highlights global sensitivity trends, and avoids inflating the narrative with redundant per-attack subsections.

The resulting table shows the weighted F1-Score for each perturbation, along with the relative variation from the baseline. In addition, attacks operating in weight space report key behavioral effects (e.g., proportion of samples redirected to the target class), since in this domain such qualitative shifts can be more informative than the raw F1-Score alone.

\begin{table*}[!t]
\centering
\caption{Robustness of the GoEmotions model under neuron- and weight-level attacks (weighted F1-Score).}
\label{tab:goemotions_attacks}

\scriptsize
\renewcommand{\arraystretch}{1.25}
\setlength{\tabcolsep}{5pt}

\begin{tabularx}{\textwidth}{
  >{\raggedright\arraybackslash}p{2.7cm}  % Attack
  >{\raggedright\arraybackslash}X         % Config
  >{\centering\arraybackslash}p{1.7cm}    % F1
  >{\centering\arraybackslash}p{1.7cm}    % Delta
}
\hline
\textbf{Attack} & \textbf{Config} & \textbf{F1-Score (weighted)} & \textbf{$\Delta$F1-Score (\%)} \\
\hline
Baseline
& --
& 0.865
& -- \\

Global neuron silencing
& 60\% neurons silenced globally
& 0.300
& -65.3 \\

Per-class silencing
& Best run (class 1), peak F1-Score = 0.882
& 0.882
& +2.0 \\

Random noise injection
& $\epsilon = 1.5$ applied to top neurons
& 0.460
& -46.8 \\

Gaussian noise
& $\sigma = 0.9$, 60\% neurons (against its own baseline of 0.608)
& 0.581
& -4.4 \\

Logit bias
& Best stable point ($\Delta = 8$). Higher $\Delta$ collapses predictions (F1-Score$\downarrow$ to 0.192).
& 0.624
& -27.9 \\

Weight attack (balanced push)
& $\Delta = 0.2$, p = 20\%, cols = 200. Flips$\rightarrow$target = 8 (16\%). Overall predicted as target = 30\%.
& 0.788
& -8.9 \\

Weight attack (unbalanced + suppression)
& $\Delta = 0.5$, p = 25\%, suppress class 4. Flips$\rightarrow$target = 29 (58\%). Overall predicted as target = 65\%.
& 0.492
& -43.1 \\

Weight attack (bias-only)
& $\Delta = 0.6$, bias-only adjustment. Flips$\rightarrow$target = 1. Stable behavior (near-baseline).
& 0.865
& 0.0 \\

Attribution-guided bit-flip
& Most significant exponent bit of the top-$k$ weights, 2.5--20\% budget. Worst case at 2.5\%; performance never leaves the baseline neighbourhood. Measured in a single pass against its own re-run baseline of 0.951.
& 0.921
& $-3.1$ \\
\hline
\end{tabularx}
\end{table*}

Table~\ref{tab:goemotions_attacks} summarizes the behavior of the GoEmotions model under the full set of neuron- and weight-level perturbations. As expected, the baseline achieves a high weighted F1-Score of 0.865, serving as a reference for the relative degradation observed across attacks.

Neuron-level attacks show heterogeneous effects. Global neuron silencing is the most disruptive intervention: removing 60\% of neurons reduces performance to 0.300 (-65.3\%), indicating a strong reliance on distributed representations. In contrast, per-class silencing behaves differently: targeting the neurons most associated with a specific emotion can improve the overall weighted F1-Score in certain configurations (peak 0.882, +2.0\%), suggesting that the model contains redundancies or spurious correlations that can be moderated by selective ablations. Noise-based attacks produce intermediate degradation. Random noise injection yields 0.460 (-46.8\%), while Gaussian noise ($\sigma = 0.9$ over 60\% of neurons) produces only a mild drop, to 0.581 against its own evaluation baseline of 0.608 (-4.4\%), reflecting a marked robustness to smooth, zero-mean perturbations compared to discrete neuron removal.

Logit-space perturbations also exhibit notable sensitivity. The logit bias attack yields a stable $\Delta = 8$ (F1-Score = 0.624, -27.9\%), whereas larger shifts collapse predictions almost entirely. This illustrates how small directional biases in the output layer can rapidly distort emotion predictions.

Weight-space attacks produce the most varied behaviors. The balanced weight push moderately harms performance (0.788, -8.9\%), although qualitative effects such as increased redirection to the target class (16\%) are more informative than F1-Score alone. The unbalanced push with suppression is substantially more aggressive, dropping the F1-Score to 0.492 (-43.1\%) and misclassifying a large fraction (65\%) of inputs into the target class. Conversely, the bias-only variant exhibits near-baseline behavior (0.865, 0.0\%), confirming that controlled modifications to the bias vector have limited global impact unless accompanied by changes to the weight matrix.

The attribution-guided bit-flip attack departs sharply from this pattern, and the contrast is informative. Flipping the most significant exponent bit of the top-ranked weights leaves the emotion classifier essentially untouched: performance stays within three points of its baseline across the entire budget, from 2.5\% to 20\% of the selected weights, whereas the same attack drives every malware detector to the collapse floor of roughly 0.07 with as little as 2--5\% of flips. The vulnerability exposed by this attack is therefore not a generic property of Transformer classifiers under bit-level corruption, but a specific fragility of the system-call detectors. A plausible reading is that the malware models concentrate their decision on a narrow, high-magnitude set of weights that separates a comparatively homogeneous input distribution, so corrupting the extreme of that set removes the discriminative signal outright, while the emotion classifier distributes the decision more evenly across the head. The cross-domain contrast is what gives the finding its force: without a second domain to compare against, the fragility of the malware detectors could be mistaken for a generic property of the architecture, and the audit would point at the wrong target. Establishing that the exposure is task-specific is precisely the kind of conclusion the framework is built to support, and it identifies weight-space redundancy in the classification head as a concrete objective for a subsequent hardening stage.

Overall, the GoEmotions experiments reveal a pattern consistent with the malware models:
Global ablations and unstructured noise are highly damaging; selective manipulations can be either harmful or beneficial; and weight-space directional attacks provide powerful, configuration-dependent mechanisms for steering classifier behavior, with the notable exception of bit-level faults, to which only the malware detectors prove vulnerable.

\subsubsection{Representation Sensitivity: \texttt{[CLS]} versus Mean Pooling}
The framework summarizes each sequence through the \texttt{[CLS]} activation, which raises the question of how much the identified neurons depend on that choice rather than on the task itself, and whether a summary concentrated in a single position discards information distributed across the remaining tokens. To quantify this, an ablation on a balanced subset recomputes the neuron ranking from mean-pooled token representations, which aggregate the hidden states of all positions at each layer, and contrasts it with the \texttt{[CLS]}-based ranking of the same model. The comparison is carried out on GoEmotions, whose six-class setting provides a controlled reference outside the malware domain, over the full set of $9{,}216$ neurons, and agreement is measured with Spearman and Kendall rank correlations together with the overlap and Jaccard index of the top-$k$ neurons at $k=10\%$ ($k=922$). Because this is an agreement check of the same kind as the attribution comparison, the result is reported in the last row of Table~\ref{tab:attribution}, alongside it.

The two representations agree moderately rather than exactly: the rankings correlate at $\rho=0.419$ ($\tau=0.285$), and they share $32.5\%$ of their top-$k$ neurons, against the $10\%$ expected by chance at this selection size. The picture that emerges is therefore one of a shared core of salient neurons, recovered irrespective of how the sequence is summarized, surrounded by a periphery that does depend on the representation. The agreement is high enough to indicate that the \texttt{[CLS]} ranking tracks task-relevant units rather than an artefact of the pooling operator, and the random-neuron control of Section~\ref{sec:results} independently confirms their functional relevance; it is not high enough to treat the two representations as interchangeable.

This moderate divergence is consistent with the behavior observed elsewhere in the evaluation. GPT-2, the only model ranked through mean pooling out of necessity, is also the model whose ranking agrees least with gradient-based attribution and the only one that fails the random-neuron control, which suggests that averaging over positions dilutes the concentration of task-relevant signal that the \texttt{[CLS]} summary preserves in encoders. The evidence thus supports the \texttt{[CLS]} representation as an efficient and functionally meaningful choice for the encoders analyzed here, while making explicit that the resulting neuron set is representation-dependent at its margins. A full token-level study, including the attention-mediated interactions through which information is distributed across positions, falls outside the scope of this work and is identified as a direction for future research.

\subsection{Computational Cost}
Table~\ref{tab:complexity} reports the wall-clock time and peak GPU memory of the two framework stages that dominate its cost, activation extraction and probe training plus ranking, measured on the malware task. Activation extraction is the principal cost and scales with the sequence length: the 512-token encoders (BERT and DistilBERT) complete it in about 7--10 seconds, whereas the 4{,}096-token models (BigBird and Longformer) require roughly 210--245 seconds. In line with the evaluation scope set out above, these figures cover the four encoder models. Probe training and ranking are negligible in comparison, under one second for every model, and peak GPU memory stays modest, between about 0.6 and 2.1~GB. The subsequent forward-hook interventions add only a constant per-neuron overhead and require no gradients. Asymptotically, activation extraction is $O(N\,L\,H)$ for $N$ samples, $L$ layers, and hidden size $H$; each intervention adds $O(k)$ per forward pass for $k$ targeted neurons; and the compact \texttt{[CLS]} or mean-pooled representation reduces the per-sample cost by a factor proportional to the sequence length relative to a full token-level analysis. These figures confirm that the framework is lightweight and scales predictably with model size and context length.

\begin{table}[!t]
\centering
\caption{Computational cost of the two dominant framework stages on the malware task: wall-clock time and peak GPU memory for each of the four encoder models.}
\label{tab:complexity}
\setlength{\tabcolsep}{6pt}
\begin{tabularx}{\columnwidth}{@{}X c c c@{}}
\toprule
\textbf{Model} & \textbf{Extraction (s)} & \textbf{Probe (s)} & \textbf{Peak GPU (MB)} \\
\midrule
BERT       & 10.0  & 0.88 & 974 \\
DistilBERT & 6.9   & 0.58 & 618 \\
BigBird    & 208.8 & 0.84 & 2104 \\
Longformer & 244.3 & 0.60 & 2143 \\
\bottomrule
\end{tabularx}
\end{table}

\section{Discussion} \label{sec:disc}
This section discusses the proposed training-free, neuron-level intervention framework and presents experimental results on malware detection and GoEmotions.

Across tasks and model architectures, the results consistently indicate that task-relevant information is distributed across broad and overlapping sets of neurons rather than concentrated in a small number of isolated units. Consequently, neuron-level interventions tend to produce gradual performance degradation, requiring increasingly large fractions of ablated neurons to induce substantial failures. At the same time, the analysis reveals architecture- and class-dependent sensitivities, where targeted interventions on specific neuron subsets can disproportionately affect certain decision pathways. These trends provide the basis for the discussion that follows.

A key strength of the framework is that it enables neuron-level analysis that is training-free, reversible, and model-agnostic. All interventions are performed at inference time via forward hooks, leaving model parameters unchanged and allowing immediate restoration after each run. By relying on \texttt{[CLS]} representations and a unified extraction procedure, the pipeline is consistent across BERT-like encoders and domains, enabling direct cross-domain comparisons. Methodologically, the framework treats explainability as an intervention-based process: neuron importance is first estimated using a lightweight linear probe, and the same neurons are then intervened upon to verify their functional impact. This strengthens the link between attribution and functional relevance and yields quantitative robustness curves that relate performance degradation to the fraction of ablated neurons, both globally and per class. The design further emphasizes reproducibility through controlled sweeps, explicit logging of neuron indices and layer scopes, and systematic checks of baseline recovery. Moreover, to our knowledge, in the context of neuron-level robustness for cybersecurity and language within a single framework, this is one of the first pipelines to unify explainability and adversarial-style interventions to reveal concrete failure budgets.

A distinction is worth drawing at this point between what the audit reports about the models it examines and what constrains the instrument itself. The performance degradation observed under global silencing depends on the ranking criterion used to select neurons. As demonstrated in \tablename~\ref{tab:ranking-effectiveness}, alternative ranking methods produce substantially greater mean macro-F1 degradation than the probe-based ranking under otherwise identical intervention conditions. Consequently, intervention effectiveness should be interpreted as ranking-dependent rather than as an intrinsic property of model robustness. The models and datasets analyzed here are pre-existing artifacts, taken as given; the contribution of this work is the protocol that exposes their internal failure modes, enabling them to be addressed in a subsequent hardening or retraining stage. The limitations of that protocol are of a different kind, and are set out next. The current implementation focuses exclusively on \texttt{[CLS]} representations for tractability, omitting token-level structure and attention pathways that may influence certain behaviors. In addition, the reliance on a linear probe assumes approximate linear separability, and alternative attribution methods could alter neuron rankings and sensitivity profiles. From a validity perspective, the evaluation relies on publicly available checkpoints and controlled experimental settings; extending the analysis to broader datasets, additional architectures, and more diverse conditions would further strengthen these findings. \addtxt{The generality of the proposed framework must be interpreted within the scope of the present evaluation. Independence from the domain and from the architecture describes the intended design of the pipeline rather than the extent of the evidence: the core extraction, ranking, and intervention stages do not encode assumptions about a particular task, label set, or application domain, and the audited model does not require retraining. The range of potential application domains is too broad to be exhaustively evaluated through individual case studies. Instead, the present study demonstrates the applicability of the protocol across two deliberately distinct domains and five models spanning two architectural families. However, the reduced reliability of the probe-based ranking on GPT-2 shows that its effectiveness does not necessarily transfer across architectures. Extending the analysis to another domain or architecture may therefore require adapting the sequence representation and calibrating intervention magnitude, while the reliability of the resulting rankings must be validated rather than assumed.} \change{A further limitation concerns architectural coverage:}{Architectural coverage is limited in a more specific sense:} the extension to the decoder-only GPT-2 relies on a mean-pooled representation and, given its 1{,}024-token context, on a truncated view of the long system-call traces. Under these conditions, the probe-based ranking becomes less reliable for GPT-2, as evidenced both by the random-neuron control, in which random silencing is at least as damaging as the probe-guided selection, and by its lower agreement with gradient-based attribution. The combined effects of mean pooling and sequence truncation may contribute to the reduced ranking reliability observed for GPT-2. However, the present experiments do not isolate the individual effects of these factors. Evaluating alternative sequence representations, longer-context decoder models, and additional ranking methods will be necessary to determine whether the reliability observed for the encoders can also be achieved in decoder-only architectures. Notably, this limitation affects the ranking rather than the detection task itself, since GPT-2 retains competitive baseline performance.

\section{Conclusions and Future Work} \label{sec:conc}
This work presents SYNAPSE, a framework for analyzing neuron-level interpretability and robustness in Transformer-based models. By extracting per-layer \texttt{[CLS]} activations, ranking neurons globally and per class, and applying fine-grained perturbations via forward hooks, SYNAPSE provides a unified methodology to analyze how models structure task-relevant information and respond to controlled interventions. Experimental results reveal a \change{consistent, domain-independent trend}{trend that holds consistently across both evaluated domains}: Transformer models distribute information across broad, overlapping neuron subsets rather than concentrating it in isolated units, introducing internal redundancy that makes neuron-level attacks more costly than expected. At the same time, the framework reveals structural weaknesses in each architecture, where certain decision pathways are disproportionately sensitive to targeted perturbations, indicating non-uniform internal organization and a reliance on narrow activation patterns. Complementary manipulations in logit space and weight space further show that small, structured changes can redirect predictions with minimal global degradation, highlighting attack surfaces not captured by neuron-centric analysis alone. Overall, the findings show that SYNAPSE supports interpretable inspection of internal representations and acts as a practical tool for stress-testing robustness in both cybersecurity and NLP applications.

Future work may extend SYNAPSE by developing neuron-level defense mechanisms that detect or mitigate targeted manipulations in realistic, security-critical deployments. Additional directions include exploring multimodal and cross-domain scenarios, where neuron attribution patterns may differ across textual, behavioral, and sensory inputs, and integrating interactive visualization tools to improve accessibility. Adapting SYNAPSE's perturbation strategies to generative and code-optimization settings, where internal representations play a central role, is another promising direction~\cite{codeoptdualencoder}. Finally, deploying the framework in federated or distributed learning settings could provide insights into neuron-level robustness when models are trained collaboratively across heterogeneous devices and threat landscapes.

\section*{Acknowledgment}
This work was supported by (a) MCIN\allowbreak/AEI\allowbreak/10.13039\allowbreak/501100011033\allowbreak/FEDER under grant PID2021-122466OB-I00, and (b) the Swiss Federal Office for Defense Procurement (armasuisse) with the TITAN project.

\bibliographystyle{elsarticle-num}
\bibliography{refs}
\end{document}